\documentclass[letterpaper]{article} 
\usepackage{aaai2026}  
\usepackage{times}  
\usepackage{helvet}  
\usepackage{courier}  
\usepackage[hyphens]{url}  
\usepackage{graphicx} 
\usepackage{natbib}  
\usepackage{caption} 
\usepackage{algorithm}
\usepackage{algorithmic}

\usepackage{newfloat}
\usepackage{listings}
\DeclareCaptionStyle{ruled}{labelfont=normalfont,labelsep=colon,strut=off} 
\floatstyle{ruled}
\newfloat{listing}{tb}{lst}{}
\floatname{listing}{Listing}
\title{UniC-Lift: Unified 3D Instance Segmentation via Contrastive Learning}
\author{
    Ankit Dhiman\equalcontrib\textsuperscript{\rm 1,\rm2},
    Srinath R\equalcontrib\textsuperscript{\rm 1},
    Jaswanth Reddy\equalcontrib\textsuperscript{\rm 1},\\
    Lokesh R Boregowda\textsuperscript{\rm 2},
    Venkatesh Babu Radhakrishnan\textsuperscript{\rm 1}
}
\affiliations{
   \textsuperscript{\rm 1}Indian Institute of Science, Bangalore \\ 
    \textsuperscript{\rm 2}Samsung R\&D Institute India - Bangalore\\
}

\usepackage{bibentry}

\usepackage{amsmath}
\usepackage{booktabs}
\usepackage{adjustbox}
\usepackage{multirow}
\usepackage{amssymb}
\usepackage{makecell}
\usepackage{bbold}

\newcommand{\methodname}{UniC-Lift}
\newcommand{\bluetick}{{\color{blue}\checkmark}}
\newcommand{\redcross}{{\color{red}$\times$}}

\newcommand{\covarmatrix}{\boldsymbol{\Sigma}}
\newcommand{\meanG}{\boldsymbol{\mu}}
\newcommand{\vectorG}{\boldsymbol{v}}
\newcommand{\rendervectorG}{\boldsymbol{\mathcal{V}}}
\newcommand{\vectormap}{\boldsymbol{\mathbb{V}}}

\begin{document}

\makeatletter
\def\@oddfoot{\hfil\thepage\hfil}
\def\@evenfoot{\hfil\thepage\hfil}
\makeatother

\maketitle

\thispagestyle{plain}

\begin{abstract}
3D Gaussian Splatting (3DGS) and Neural Radiance Fields (NeRF) have advanced novel-view synthesis. Recent methods extend multi-view 2D segmentation to 3D, enabling instance/semantic segmentation for better scene understanding. A key challenge is the inconsistency of 2D instance labels across views, leading to poor 3D predictions. Existing methods use a two-stage approach in which some rely on contrastive learning with hyperparameter-sensitive clustering, while others preprocess labels for consistency. We propose a unified framework that merges these steps, reducing training time and improving performance by introducing a learnable feature embedding for segmentation in Gaussian primitives. This embedding is then efficiently decoded into instance labels through a novel "Embedding-to-Label" process, effectively integrating the optimization. While this unified framework offers substantial benefits, we observed artifacts at the object boundaries.
To address the object boundary issues, we propose hard-mining samples along these boundaries. However, directly applying hard mining to the feature embeddings proved unstable. Therefore, we apply a linear layer to the rasterized feature embeddings before calculating the triplet loss, which stabilizes training and significantly improves performance. Our method outperforms baselines qualitatively and quantitatively on the ScanNet, Replica3D, and Messy-Rooms datasets.
\end{abstract}

\begin{links}
    \link{Project Page}{https://unic-lift.github.io/}
    \link{Code}{https://github.com/val-iisc/UniC-Lift}
\end{links}

\section{Introduction}
\label{sec:intro}

\begin{figure}[!t]
    \centering
    \includegraphics[width=\linewidth]{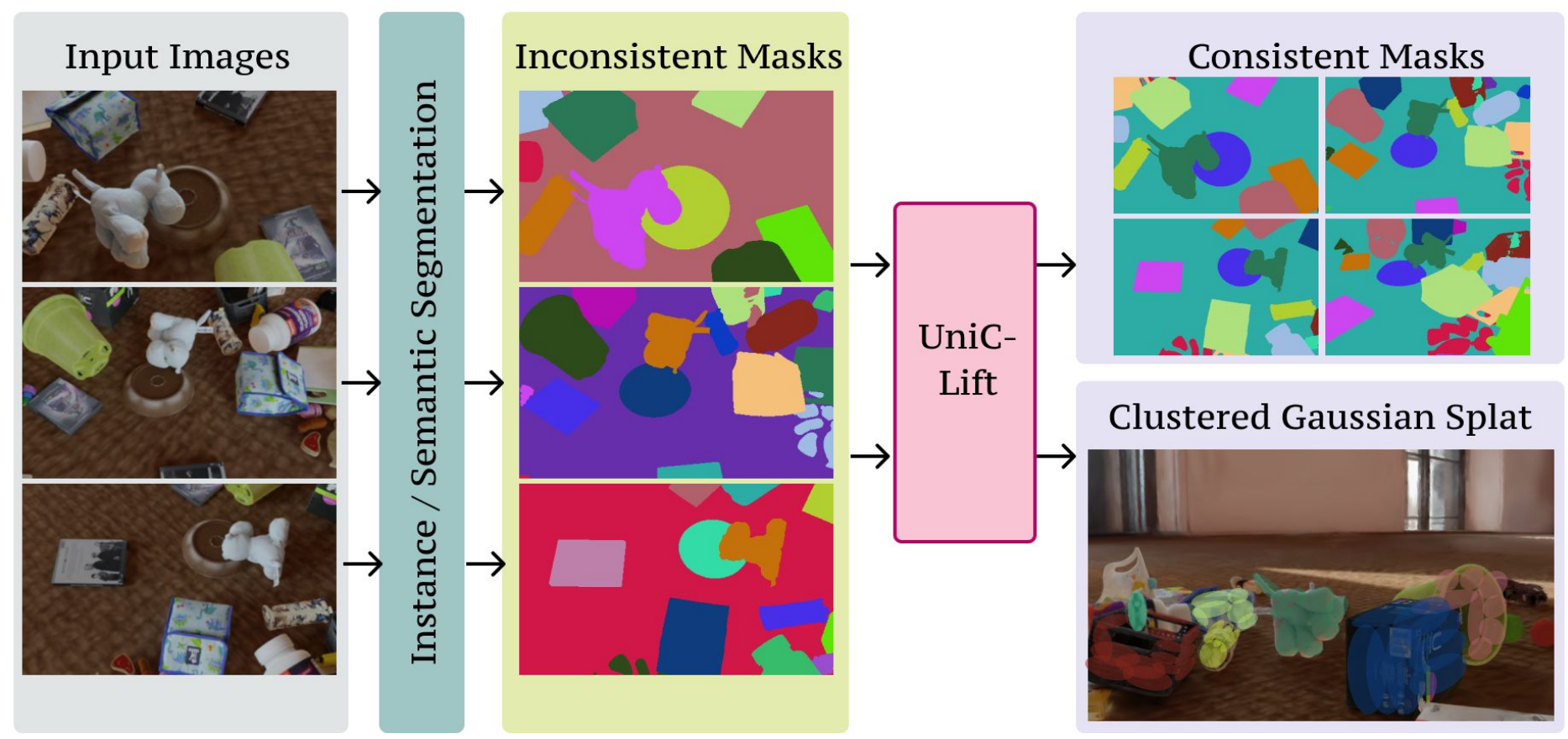}
    \caption{Our method takes a set of multi-view RGB images as input, and passes these images through a pre-trained instance/semantic segmentation method. Although accurate, segmentation methods for 2D images do not generate multi-view consistent segmentation masks. Observe how labels vary across the images. Our method formulates this as a clustering problem in 3D, utilizing feature-vectors in 3D space which are trained using contrastive losses. Our method generates consistent segmentation masks. We further motivate the necessity of this work in Fig.~\ref{fig:why_fig}}
    \label{fig:teaser}
\end{figure}

Extensive research has been undertaken on various 2D segmentation variants, including semantic~\cite{xu2023pidnet, zhou2023zegclip}, instance~\cite{he2023fastinst, cheng2022masked}, and panoptic~\cite{Kirillov2019PanopticFP, Hu2023YouOS} segmentation. Recently, advancements in open-set segmentation~\cite{kirillov2023segment, zou2024segment} have enabled these techniques to predict classes absent in the training dataset. Building upon these 2D advancements, researchers have focused on understanding 3D scenes, which is crucial for applications in AR/VR~\cite{Choy20194DSC}, autonomous driving~\cite{Feng2019DeepMO}, and path planning~\cite{Bartolomei2020PerceptionawarePP}. A major challenge in 3D scene understanding is the lack of large-scale, richly annotated datasets, unlike 2D image datasets (e.g., Cityscapes~\cite{Cordts2016Cityscapes}, MS COCO~\cite{lin2015microsoft}, CamVid~\cite{brostow2009semantic}). To overcome this, many works ``lift'' labels from 2D segmentation methods to 3D representations such as point clouds~\cite{Liu2022PartSLIPLP}, Neural Radiance Fields (NeRF)~\cite{kundu2022panoptic, siddiqui2023panoptic}, and Gaussian Splatting (3DGS)~\cite{zhou2023feature}. This approach uses efficient 2D segmentation techniques for 3D tasks but faces a key challenge: \textit{inconsistent multi-view semantic masks from 2D models}.

In this work, we tackle the problem of 3D instance segmentation under inconsistent 2D instance masks. (Fig.~\ref{fig:teaser}). One plausible solution for the task is to use an off-the-shelf video instance segmentation method~\cite{Zhang2023BoostingVO, heo2023generalized} to obtain consistent masks for multi-view image sequences. However, these methods have limitations when there are significant occlusions and large changes in the observed scene, thus requiring joint optimization for re-identification task~\cite{Li2018VideoOS}, which is challenging for real-world in-the-wild scenes. Recent 3D representations (NeRF, 3DGS) learn the underlying geometry of the scene from 2D multi-view images. Thus, these representations can be utilized to aggregate semantic/instance labels from 2D segmentation methods and generate 3D consistent masks. Recent works ~\cite{fu2022panoptic, kundu2022panoptic, wang2022dm, zhi2021place, siddiqui2023panoptic} in this direction lift the 2D segmentation labels to 3D representations to achieve multi-view consistent 3D segmentation. However, a significant limitation of these methods is that they are extremely slow as they have additional expensive steps of label assignment~\cite{siddiqui2023panoptic} in the loss function.

Another promising line of work~\cite{kobayashi2022decomposing, zhou2023feature} is to distill features from rich 2D image feature extractors such as CLIP-LSeg~\cite{li2022language} or DINO~\cite{caron2021emerging} instead of lifting the 2D labels. However, these methods suffer from slow speed as they distill high-dimensional features into 3D representations. For example, on average, DFF~\cite{kobayashi2022decomposing} takes approximately two days of training.  
Recently, Contrastive-Lift~\cite{bhalgat2023contrastive} solved this problem by connecting 2D segmentation to 3D using a learnable 3D vector embedding that aligns with 2D label predictions. Contrastive-Lift achieves this by integrating slow-fast contrastive loss, eliminating the need to solve the linear assignment problem. However, this method requires an additional expensive step of HDBSCAN~\cite{mcinnes2017hdbscan} as post-processing to find the centroids to label the optimized embedding. In this work, we propose an efficient \textit{single-stage} 3D segmentation method achieving superior performance in a fraction of training time compared to baseline.

We propose~\emph{\methodname}, a novel 3D segmentation method built on 3DGS~\cite{kerbl20233d}. ~\emph{\methodname} addresses the challenge of generating consistent 3D segmentation masks from inconsistent 2D segmentation maps. To effectively encode 3D features, we add a d-dimensional vector embedding $v\in{\mathbb{R}}^{d}$ as a property for each 3D Gaussian primitive and then rasterize this embedding to the desired camera view. 
Further, to address the issues at the boundary, we propose a strategy for hard-mining samples to apply the triplet loss.  
We propose a simple ``Embedding-to-label'' process to decode the rasterized embeddings to class label.
This straightforward approach outperforms existing state-of-the-art 3D segmentation methods. We demonstrate the utility of~\emph{\methodname} in two downstream tasks: object manipulation and object extraction. Our main contributions are as follows:
\begin{itemize}
    \item An effective single-stage method for 3D segmentation that directly decodes the learned 3D embedding to a consistent segmentation label, given inconsistent 2D segmentation labels.

    \item A novel contrastive loss based on the triplets to reduce the inter-class variance which aids in accurate 3D segmentation.  

    \item Demonstrating high-quality and accurate 3D segmentation by showing effective downstream object manipulation and extraction applications. 
 
\end{itemize}

\section{Related Work}

\noindent
\textbf{3D Representations.} Neural Radiance Fields~\cite{mildenhall2021nerf} and its variants~\cite{barron2022mip, muller2022instant, barron2021mip, fridovich2022plenoxels, chen2022tensorf, barron2023zip} utilize volumetric rendering equation which is based on ray-tracing to make a differentiable renderer. These representations perform well on the task of novel-view synthesis from input multi-view images and their camera pose. Recently, Gaussian Splatting~\cite{kerbl20233d} has revolutionized this field. It is based on rasterization instead of ray-tracing and can achieve very high FPS to render high-resolution novel-views~\cite{lu2024turbogsaccelerating3dgaussian}. Further, these representations are extended to other tasks such as dynamic scenes~\cite{ yang2023deformable3dgs}, stylization~\cite{huang2022stylizednerf, dhiman2025chromadistill}, sparse-views~\cite{li2024dngaussian, wu2023reconfusion}, hierarchical scenes~\cite{dhiman2023strata}, etc.

\noindent
\textbf{Scene Understanding in 3D Representations.} Semantic-NeRF~\cite{zhi2021place} lifts 2D semantic labels to 3D by using a separate radiance field network to predict the class labels utilizing a cross-entropy based loss. Whereas NeSF~\cite{vora2021nesf} samples a pre-trained NeRF model to obtain the volumetric grid and convert this to a semantic grid by using a convolutional volume-to-volume network. This semantic grid is converted into class probabilities by rendering. These methods did not take into account the inconsistency in the 2D ground-truth semantic maps. Afterwards, a lot of works~\cite{mirzaei2022laterf, yu2021unsupervised} explored the similar approach and improved the performance in scene understanding using NeRF. Similar techniques~\cite{lan20232d, li2024sgs} have also been explored in the domain of Gaussian splatting. 

Methods such as Panoptic NeRF~\cite{fu2022panoptic} and Instance-NeRF~\cite{liu2023instance} use information from 3D instances by extracting volume density from the pre-trained NeRF network. The major disadvantage of these works was the use of 3D masks or tracked object masks. Panoptic Lifting~\cite{siddiqui2023panoptic} and DM-NeRF~\cite{wang2022dm} use linear assignment problem during optimization to solve the multi-view inconsistency problem in the ground-truth 2D mask. Contrastive-Lift~\cite{bhalgat2023contrastive} used learnable permutation-invariant embedding vectors to solve the multi-view inconsistency problem. But rely on algorithms like HDBSCAN to find the cluster centroids. 

\noindent
\textbf{Open-Set segmentation in 3D.} Recently, methods such as SAM~\cite{kirillov2023segany}, SEEM~\cite{zou2024segment} which has capability of segmenting objects in a scene by using texts or interactive scribbles. DFF~\cite{kobayashi2022decomposing} distills the knowledge of off-the-shelf 2D image feature extractors such as CLIP-LSeg~\cite{li2022language} or DINO~\cite{caron2021emerging} into a 3D feature field optimized in parallel to the radiance field. Then, these distilled features are used to decompose the 3D scene and perform various editing tasks. Based on the similar strategy, other works~\cite{tschernezki2022neural, kerr2023lerf} distill open-set semantics into 3D space which can then be used for zero-shot segmentation. Similar to NeRF. a lot of work~\cite{shi2023language, qin2023langsplat, zhou2023feature} have also emerged for Gaussian splitting.

\begin{figure}[!t]
  \centering
  \includegraphics[width=\linewidth]{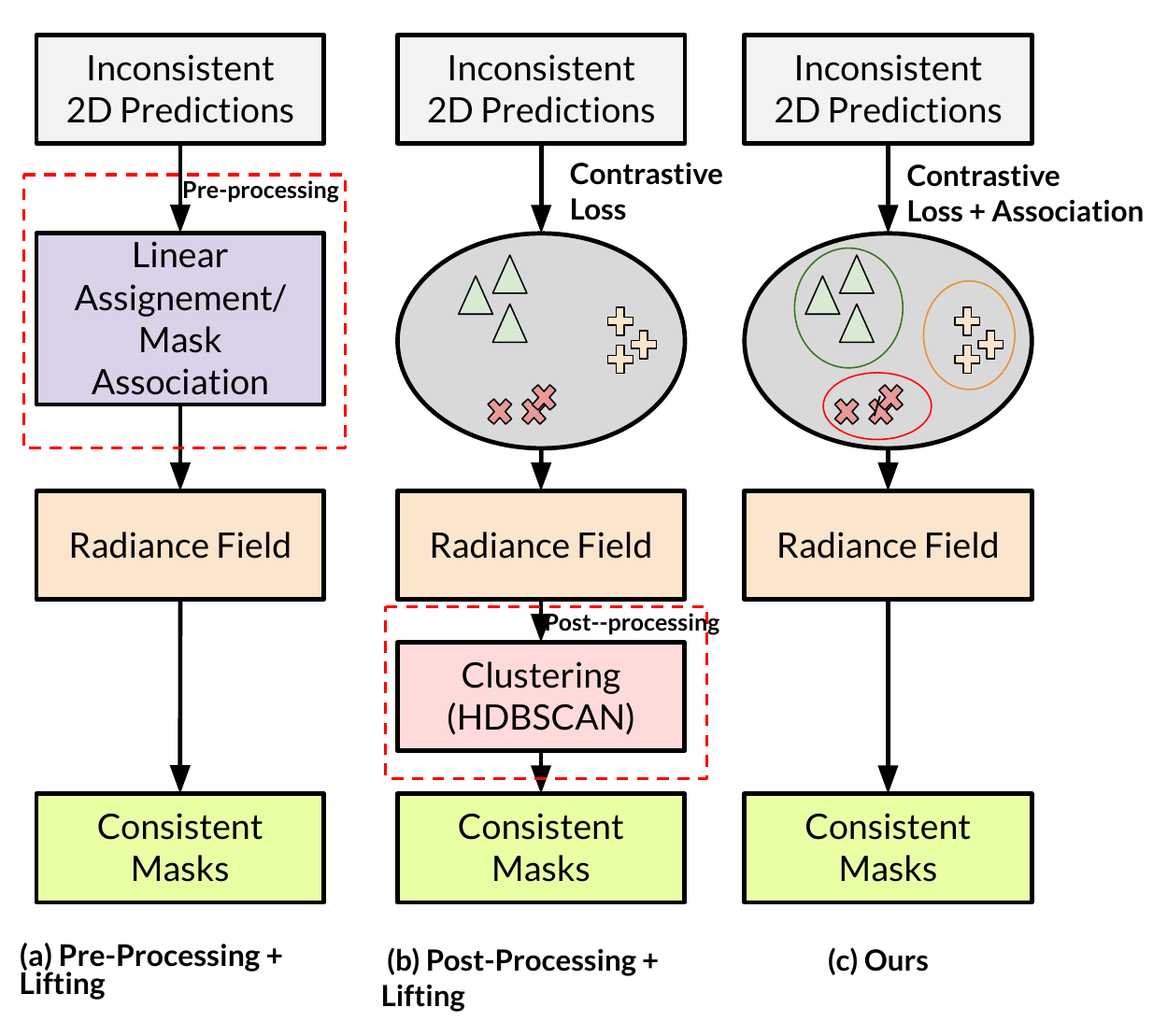} 
  \caption{
  Previous methods are multi-stage: (a) some pre-process 2D masks before lifting them to 3D (e.g., DM-NeRF~\cite{wang2022dm}, Gaussian Grouping~\cite{ye2023gaussian}, and Panoptic-Lift~\cite{kundu2022panoptic}), while others use (b) a separate clustering algorithms on learned embeddings (e.g., Contrastive-Lift~\cite{bhalgat2023contrastive}). (c) In contrast, our method uses a single, unified representation to perform segmentation directly.
  }
  \label{fig:why_fig}
\end{figure}

\section{Proposed Method}
\label{sec:method}

\subsection{Preliminaries}
\label{subsec:preliminaries}

3D Gaussian Splatting (3DGS) models a scene using set of 3D Gaussians parameterized by position ($\meanG$), covariance ($\covarmatrix$), opacity ($\alpha$), and spherical harmonics for color ($c$). Given input RGB images and camera poses, 3DGS jointly optimizes these parameters to render novel views. Gaussians are initialized from  SfM point cloud and refined via ``Adaptive Density Control'' algorithm which splits, clones, and prunes primitives based on training gradients. 
During rendering, Gaussians are projected to $T$ 2D splats, rasterized, and $\alpha$-blended to get the final pixel color $C$.
\begin{equation}
    C = \sum_{i \in T} c_i\alpha'_i\prod_{j=1}^{i-1}(1 - \alpha'_{j}). \label{eq:color}
\end{equation} 
The final opacity $\alpha'_i$ is determined by multiplying the learned opacity $\alpha_i$ with the 2D projected Gaussian density evaluated at the pixel location. The rendering loss function is a combination of $\mathcal{L}_1$ and D-SSIM loss:
\begin{equation}
\mathcal{L}_{rendering} = (1 - \lambda)\mathcal{L}_1 + \lambda\mathcal{L}_{ssim}  
     \label{eq:rendering_loss}
\end{equation}

\begin{figure}[!t]
    \centering
    \includegraphics[width=\linewidth]{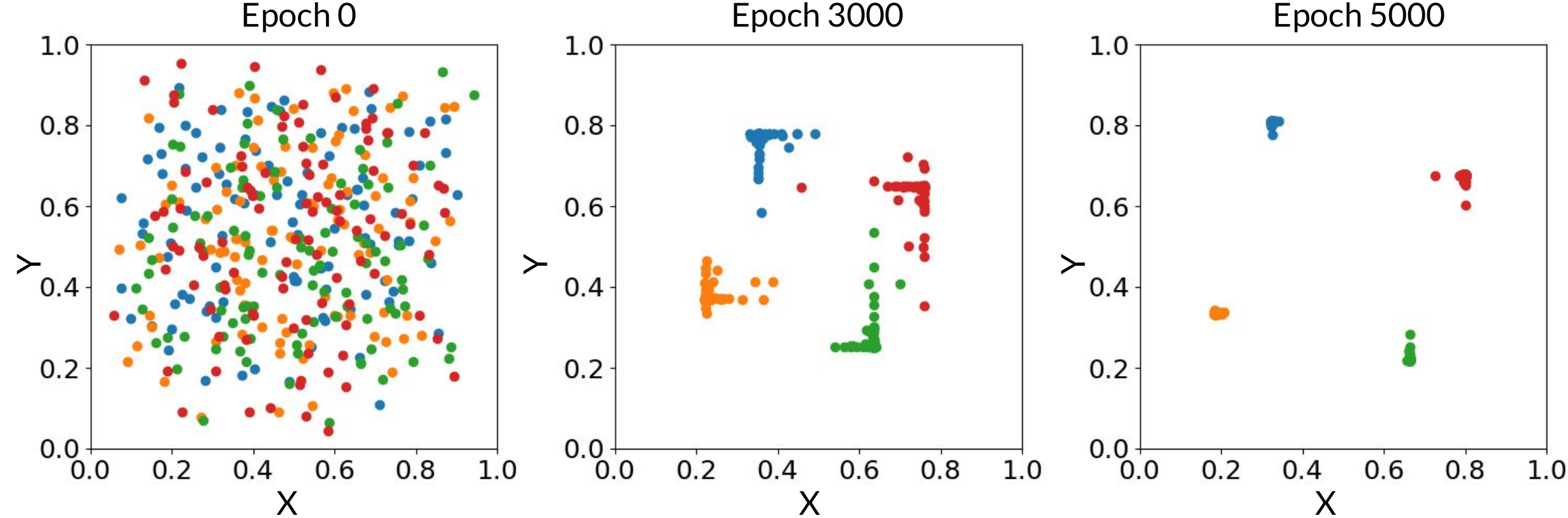}
    \caption{\textbf{Intuition of Embedding-To-Label Process.} A toy experiment demonstrating how contrastive learning enables direct label prediction. Embeddings, initially distributed in a constrained space (e.g., $[0,1]^2$), converge to distinct corners during training. Each corner effectively becomes a unique binary code that can be thresholded and mapped to a label, removing the need for post-processing.}
    \label{fig:rebuttal-embedding-to-label}
\end{figure}

\subsection{Motivation}

Two-stage methods like Contrastive-Lift require a computationally expensive post-processing step, using algorithms such as HDBSCAN to derive labels from optimized embeddings. This approach introduces significant computational overhead during training (Tab.~\ref{tab:discussion_hdbscan}). Further, it also results in a complexity for novel views of $O(n\,log\,{c})$, where $n$ is the number of pixels and $c$ is the number of clusters. In contrast, our unified, single-stage framework eliminates this bottleneck by directly associating labels with the optimized vectors. This design achieves a superior $O(n)$ time complexity for novel-view prediction, reducing overall inference time, and represents a fundamental architectural improvement over prior methods, as illustrated in Fig.~\ref{fig:why_fig}.

\begin{figure*}[!t]
    \centering
    \includegraphics[width=0.93\linewidth]{Figures/architecture_tight.pdf}
    \caption{Overview of our pipeline. Given input (a) multi-view RGB images and (c) inconsistent segmentation maps, we optimize (b) 3D representation to lift 2D segmentation labels to 3D. To learn the view-dependent changes associated with the novel-view synthesis task, we take a rendering loss: $\mathcal{L}_{rendering}$ with GT RGB image. For optimizing the learnable 3D vector embedding we apply $\mathcal{L}_{cluster}$ to the rasterized embedding and further apply $\mathcal{L}_{triplet}$ by passing it through a linear layer. Further, we apply a 3D loss $\mathcal{L}_{3D}$ to the primitives of 3D representation. More details are in Method section. }
    \label{fig:architecture}
\end{figure*}

\textbf{Embedding-To-Label Process.}
A key challenge in developing a unified contrastive learning framework is the decoding of optimized embeddings into discrete labels. We illustrate the intuition behind our approach by presenting a simplified toy experiment that demonstrates how our learned embeddings are transformed into discrete labels in a unified, single-stage process.
\underline{\textbf{i. Experimental Setup:}} Consider a scenario where we train a model using contrastive learning on embeddings constrained to the $[0,1 ]^2$ space via a sigmoid activation. The goal is to cluster these embeddings into at least four distinct classes. \underline{\textbf{ii. Observation}:} Initially, the embedded vectors are randomly distributed within the square as shown in Fig.~\ref{fig:rebuttal-embedding-to-label}. Embeddings of similar instances are pulled together and different instances are pulled apart due to the applied contrastive clustering loss.As training proceeds, the embeddings align and converge towards the corners of the square, forming well-separated clusters. \underline{\textbf{iii. Conversion to Label:}} The conversion to discrete labels is straightforward. For e.g, an embedding clustered near the top-right corner (e.g., $[0.9, 0.8]$), after thresholding by 0.5, yields a binary vector $[1, 1]$ and decodes to 3. Each binary vector directly decodes to a specific label (e.g., [1,1] decodes to 3, with other corners decoding to 0, 1, and 2).

\subsection{Methodology}
\label{subsec:methodology}
As explained in Preliminaries, a 3D Gaussian is parametrized by $\meanG$, $\covarmatrix$, $\alpha$, and color $c$ in 3DGS. We add an extra parameter, vector $\vectorG$, where $\vectorG \in {\mathbb{R}}^{d}$ is a d-dimensional vector to model instance segmentation of a scene. Given set of images $\mathcal{I} = \{I_1, I_2, ..., I_N\}$, their corresponding poses $\pi = \{{\pi}_1, {\pi}_2, ..., {\pi}_N\}$ and the semantics or instance segmentation labels $\mathcal{M} = \{M_1, M_2, ..., M_N\}$, our method also optimizes for the vector $v$ along with other parameters of 3DGS. Note that $M_i \in {\mathbb{R}}^{H \times W}$,  $I_i \in {\mathbb{R}}^{H \times W \times 3}$ where $H$ and $W$ are the height and width of an input image.
We render $\vectorG$ to get vector-embedding $\rendervectorG$, on which we apply contrastive loss and then pass it through a linear layer. Then, we apply triplet loss between positive and negative pairs along the boundaries of the 2D segmentation map. Further, we regularize 3D Gaussians with a smoothness loss such that neighbouring Gaussians have a similar vector $\vectorG$. During inference, we apply a simple ``Embedding-to-Label'' process to the rendered vector $\rendervectorG$ to get the class label.   

\noindent
\textbf{Rendering of learnable vector-embeddings.} 
Each 3D Gaussian has a vector embedding $\vectorG$ for instance or semantic segmentation. Unlike color, $\vectorG$ is view-independent. During training, we initialize $\vectorG$ using a normal distribution and render these vector-embeddings in the same way as color.
\begin{equation}
    \rendervectorG = \sum_{i \in T} \vectorG_i\alpha'_i\prod_{j=1}^{i-1}(1 - \alpha'_{j})
\end{equation}

\begin{figure*}[!t]
  \centering
  \includegraphics[width=\textwidth]{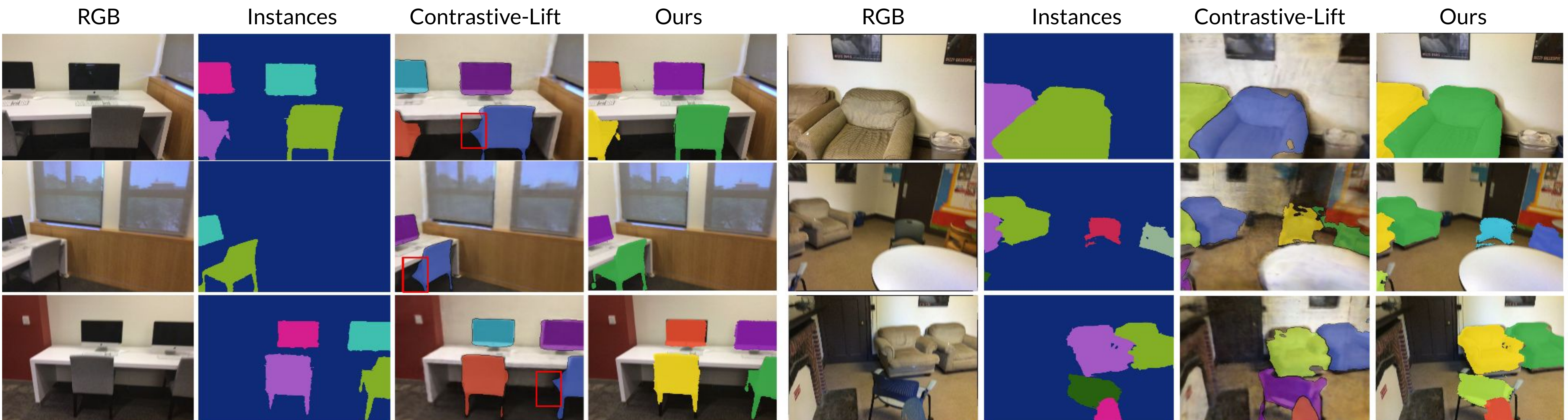}
  \caption{\textbf{Qualitative comparison of our method with Contrastive lift on scenes from ScanNet dataset.}  Regions, where Contrastive lift performs poorly, are highlighted with red boxes.}
  \label{fig:qualitative_comparison}
\end{figure*}

\noindent
\textbf{Contrastive Loss on rendered vector-embeddings 
($\rendervectorG$).} 
Building upon the rendered vector-embeddings $\rendervectorG$, we generate a vector map $\vectormap \in {\mathbb{R}}^{H \times W \times d}$ for a given camera pose $\pi_{i}$. Corresponding to the camera pose, we also have a ground truth segmentation mask $M_i$. Using $M_i$, we divide the pixel space into $K$ distinct sets, each representing a segment. This creates a partition $\mathcal{O}=\{\Omega_1, \Omega_2, ..., \Omega_K\}$, where each $\Omega_i \cap \Omega_j = \emptyset$ is disjoint from the others. Next, we compute the centroid for each segment as $\boldsymbol{ m_{\Omega_i} } = \frac{1}{|\Omega_i|} \sum_{u \in \Omega_i} \vectormap(u) $. The core idea is to apply a contrastive loss that maximizes the similarity between vector-embeddings within the same segment and minimizes the similarity between vector-embeddings from different segments. This clustering loss is then calculated as follows:

\begin{equation}
    \begin{split}
    \mathcal{L}_{\text{cluster}} &= \sum_{\Omega_i \in \mathcal{O}} 
    \sum_{u \in \Omega_i}\left\| \vectormap(u) - \boldsymbol{ m_{\Omega_i} } \right\|_2^2 - \\
    &\sum_{\substack{\Omega_i, \Omega_j \in \mathcal{O} \\ i \neq j}}
    \left\| \boldsymbol{ m_{\Omega_i} } - \boldsymbol{ m_{\Omega_j} } \right\|_2^2
\end{split}
\label{eq:cluster}
\end{equation}

\noindent
\textbf{Triplet Loss.} 
To cluster the vector embeddings $\rendervectorG$, we  use the clustering loss that measures the similarity between two vector-embeddings. However, this distance metric does not guarantee a consistent penalty, as detailed in the supplementary material. To address this, we transform $\vectormap$ by passing it through a linear layer, parametrized by $\mathbb{W}^{d \times d}$, resulting in a transformed vector $\boldsymbol{z} =\mathbb{W}\rendervectorG$. Before passing through linear layer, we pass $\rendervectorG$ through a sigmoid layer to restrict the range to $[0,1]$.  Instead of clustering $\boldsymbol{z}$ directly, we apply a triplet loss. This loss further reduces variance within clusters and maximizes the margin between different clusters. Our triplet sampling strategy is straightforward. For each pixel $u_1,u_2\in\Omega_i$, we select vector-embedding $a=\mathbb{W}\,\sigma(\mathbb{V}(u_1))$ as an anchor. We then randomly choose a positive sample $p=\mathbb{W}\,\sigma(\mathbb{V}(u_2))$ and a negative sample $n=\mathbb{W}\,\sigma(\mathbb{V}(u_3))$ from a different segment $u_3\in\Omega_j$, where $i\neq j$ and $\sigma(.)$ is sigmoid operator. To enhance cluster separation, we select positive and negative samples from segment boundaries. After iterating through all pixels, we obtain a set of triplets $\Delta$, which we use to enforce the triplet loss as defined in Eq.~\ref{eq:triplet} where $\delta$ is the margin. Note that the triplets are obtained in the projected space through linear layer mentioned earlier.

\begin{equation}
 \mathcal{L}_{triplet} = \sum_{(a,p,n) \in \Delta} max(0,\mathcal||a - p||_2^{2} - \mathcal||a - n||_2^{2} + \delta)
\label{eq:triplet}
\end{equation}

\noindent
\textbf{3D neighborhood Regularization.} 
To encourage nearby Gaussians to share similar embeddings, we 
apply a spatial smoothness prior in the 3D space.
For each Gaussian primitive $i$ with center $\mu_i$, we construct a 3D neighborhood set $\mathcal{N}(i) = \{\|\mu_i - \mu_j\|_2^2 \leq \tau; j \neq i; \; i,j \in \{1,...,|\mathcal{G}| \} \}$, where $\tau$ is a threshold which selects only those 
Gaussians that are physically close and $|\mathcal{G}|$ are the number of Gaussian primitives in 3DGS model $\mathcal{G}$.
We then enforce embedding consistency only within these neighborhoods by penalizing the difference between their embedding vectors:
\begin{equation}
  \mathcal{L}_{3D} = \sum_{i=1}^{|\mathcal{G}|} \sum_{\substack{j\in \mathcal{N}(i)}}
\|\vectorG_i - \vectorG_j\|_2^2
 \label{eq:3d_loss}
\end{equation}
We apply this loss after $15000$ iterations, once the adaptive density control mechanism has stabilized.

\noindent
Total loss function for this optimization problem is:
\begin{equation}
\begin{split}
 \mathcal{L}_{total} = \mathcal{L}_{rendering} + \lambda_{cluster} \mathcal{L}_{cluster} \\ +\lambda_{triplet} \mathcal{L}_{triplet} + \lambda_{3D} \mathcal{L}_{3D}
 \label{eq:tot_loss}
 \end{split}
\end{equation}

For better stability, we apply $\mathcal{L}_{triplet}$ and $\mathcal{L}_{3D}$ after the adaptive density control step. We use $\lambda_{cluster}$, $\lambda_{triplet}$, $\lambda_{3D}$, $\delta$ and $\tau$ as $1e^{-1}$, $1e^{-1}$, $1e^{-1}$, $1$,  and $1e^{-2}$, respectively.

\noindent
\textbf{Embedding-to-label.} 
We first convert the rendered embedding $\rendervectorG$ by applying a sigmoid function $\hat{\rendervectorG}=\sigma(\rendervectorG)$. We then threshold it to convert into a binary vector $\tilde{\rendervectorG}=1[\hat{\rendervectorG}>\tau]$, where $\tau=0.5$. We obtain discrete label by mapping $\tilde{\rendervectorG}$ into a label $l=\sum_{k} \tilde{\rendervectorG}_k\, 2^{\,k-1}$. Further details and rationale are provided in the supplementary material.

\section{Experiments}
\label{sec:exp}

\begin{table*}[!t]
\centering

\begin{adjustbox}{width=0.98\linewidth}
\begin{tabular}{lccccccccc}
\toprule
\multirow{2}{*}{Method} & \multicolumn{4}{c}{Old Room environment} & \multicolumn{4}{c}{Large Corridor environment} & Mean (\%) \\
\cmidrule(l{4em}r{4em}){2-5} \cmidrule(l{4em}r{4em}){6-9}
& 25 Objects & 50 Objects & 100 Objects & 500 Objects & 25 Objects & 50 Objects & 100 Objects & 500 Objects &  \\
\midrule
Panoptic-Lifting~\cite{siddiqui2023panoptic}  & 73.2 & 69.9 & 64.3 & 51.0 & 65.5 & 71.0 & 61.8 & 49.0 & 63.2\\
Contrastive-Lift~\cite{bhalgat2023contrastive} & 78.9 & 75.8 & 69.1 & 55.0 & 76.5  & 75.5  &  68.7  & 52.5 & 69.0 \\
OmniSeg3D-GS~\cite{ying2024omniseg3d} & 80.1 & 72.4 & 61.4 & 46.8 & 74.9 & \textbf{79.6}  &  63.9  & 48.5 & 66.0\\
Unified-Lift~\cite{zhu2025rethinking} & 79.1 & 72.2 & 65.9 & 53.9 & \textbf{77.0} & 78.9  &  70.7  & 54.1 & 69.0\\
\textbf{Ours} & \textbf{86.0} & \textbf{79.1} & \textbf{70.8} & \textbf{57.4} & 76.5  & 72.3  & \textbf{73.0}  & \textbf{56.7} & \textbf{71.5}\\
\bottomrule
\end{tabular}
\end{adjustbox}
\caption{\textbf{Messy-Rooms dataset.} PQ$^\text{scene}$ metric is reported and best results are marked in \textbf{bold}. Results for the baseline methods are sourced from~\cite{zhu2025rethinking}. We observe that our method outperforms the baseline methods in $6$ out of $8$ scenes.}
\label{tab:messy_rooms_quant}
\end{table*}

\noindent
\textbf{Implementation Details}
We build our method on the 3DGS~\cite{kerbl20233d} and train it for $30k$ iterations on a single RTX A6000 GPU. The model is optimized using the ADAM optimizer~\cite{kingma2014adam} with a learning rate of $1e^{-4}$ and the loss defined in Eq.~\ref{eq:tot_loss}. We exclude gradients from the segmentation losses (Eqs.~\ref{eq:cluster},~\ref{eq:triplet},~\ref{eq:3d_loss}) from the 3DGS ``Adaptive Density Control'' step. For our experiments, we set the embedding dimension ($d$) to $12$ and use a maximum of $3,000$ triplets.

\noindent
\textbf{Dataset and Baselines}
 We benchmark our results on ScanNet~\cite{dai2017scannet} and Replica3D~\cite{straub2019replica} datasets using the same scenes as Contrastive-Lift. We also evaluate on Messy-Rooms dataset, which contains scenes with 25 to 500 objects. We compare our method with Contrastive-Lift~\cite{bhalgat2023contrastive}, Panoptic-Lifting~\cite{fu2022panoptic}, Panoptic-NeRF~\cite{kundu2022panoptic}, Gaussian-Grouping~\cite{ye2023gaussian} and Unified-Lift~\cite{zhu2025rethinking}. We follow the Contrastive-Lift evaluation protocol for ScanNet and Replica. 

\noindent
\textbf{Metrics.} We evaluate label accuracy using mean intersection over Union (mIoU) and label consistency using scene-level Panoptic Quality (PQ$^{\text{scene}}$). PQ$^{\text{scene}}$~\cite{siddiqui2023panoptic} extends Panoptic Quality (PQ)~\cite{kirillov2019panoptic} by performing matching at the scene level: predicted and ground-truth segments of the same class are merged across views, and pairs with IoU $> 0.5$ are treated as valid matches. We additionally learn a semantic embedding per primitive alongside the instance embedding, and rasterize both to obtain instance and semantic labels to compute PQ$^{\text{scene}}$.

\begin{table*}[!t]

\begin{minipage}{0.63\linewidth}
\begin{adjustbox}{width=\linewidth}
\begin{tabular}{@{}c|ccccccc@{}}
\toprule
\textbf{Method} &
  \textbf{\begin{tabular}[c]{@{}c@{}}DM- \\ NeRF\end{tabular}} &
  \textbf{PNF} &
  \textbf{\begin{tabular}[c]{@{}c@{}}PNF \\ +GT Boxes\end{tabular}} &
  \textbf{\begin{tabular}[c]{@{}c@{}}Panoptic-\\ Lifting\end{tabular}} &
  \textbf{\begin{tabular}[c]{@{}c@{}}Contrastive-\\ Lift\end{tabular}} &
  \textbf{\begin{tabular}[c]{@{}c@{}}Gaussian-\\ Grouping\end{tabular}} &
  \textbf{Ours} \\ \midrule
\textbf{ScanNet} & 
   41.7 &
   48.3 &
   54.3 &
   58.9 &
   62.3 &
   61.83 &
   \textbf{63.0} \\ \bottomrule
\textbf{Replica3D} &
   44.1 &
   41.1 &
   52.5 &
   57.9 &
   59.1 &
   66.52 &
   \textbf{88.7} \\ \bottomrule
\end{tabular}
\end{adjustbox}
\caption{Results on ScanNet and Replica datasets. We report the PQ$^\text{scene}$ metric and source these values from~\cite{bhalgat2023contrastive}. We observe that our method outperforms baselines in both the datasets. Our method achieves a $1.3\times$ higher $PQ^{scene}$ ($88.7$ vs. $66.52$) than Gaussian Grouping.}
\label{tab:quantative_evaluation}
\end{minipage} \hfill \hfill
\begin{minipage}{0.35\linewidth}
    \begin{adjustbox}{width=\linewidth}
    \begin{tabular}{@{}c|ccc@{}}
    \toprule
    \textbf{Method} &  \begin{tabular}[c]{@{}c@{}}Panoptic\\ Lifting\end{tabular}
     &  \begin{tabular}[c]{@{}c@{}}Contrastive\\ Lift\end{tabular}
     & Ours\\ \midrule
    \textbf{Training}     &    $>20\,hrs$   &  $>15\,hrs$  &    \textbf{$<40\,mins$}  \\ \bottomrule
    \end{tabular}
    \end{adjustbox}
    
    \caption{Training time comparison on NVIDIA A6000 for a scene in Replica dataset. Our method exhibits a significant training time advantage for lifting 2D segmentation masks to 3D over other methods.}
  \label{tab:time_comparison}

\end{minipage}
\end{table*}



\noindent
\textbf{Quantitative Results} Tab.~\ref{tab:quantative_evaluation} shows quantitative comparison on ScanNet and Replica3D dataset. We follow the setup described in Contrastive-Lift. We observe that our method outperforms other methods in both datasets. Our approach shows a significant gain of nearly $10$ points for the Replica3D datasets. We evaluate our method on the Messy-Rooms dataset (Tab.~\ref{tab:messy_rooms_quant}). This dataset consists of a large number of objects, allowing us to assess the scalability of our approach. Our method outperforms Contrastive-Lift in 6 out of 8 scenes within this dataset, demonstrating its effectiveness. Further, our method also outperforms the recent method Gaussian-Grouping~\cite{ye2023gaussian} on both ScanNet and Replica3D dataset.

\begin{figure*}[!t]
  \centering
  \includegraphics[width=\textwidth]{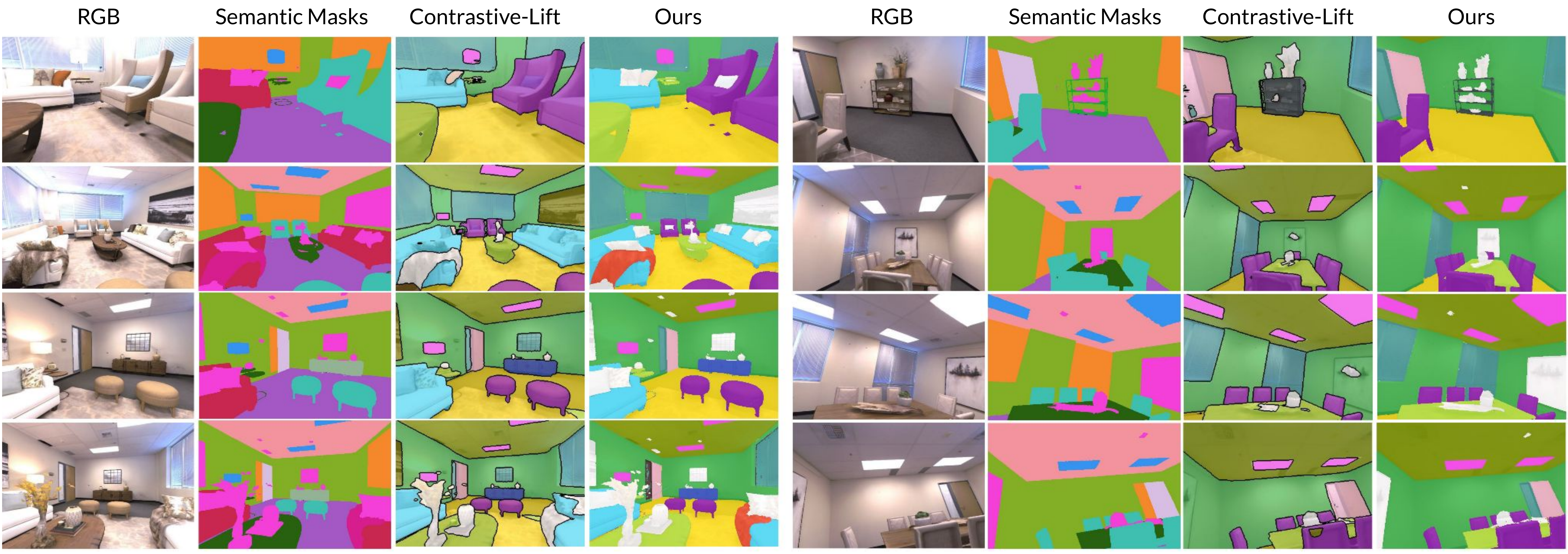}
  \caption{\textbf{Comparison with Contrastive lift on Replica3D dataset.} Observe that Contrastive-Lift fails to capture some objects such as the pillows on the sofa set and the set of objects placed inside the mini-cupboard }
  \label{fig:qualitative comparison_2}
\end{figure*}

\noindent
\textbf{Qualitative Results.} We compare our method with Contrastive-Lift in Fig.~\ref{fig:qualitative_comparison} and~\ref{fig:qualitative comparison_2}. We can observe that our method generates accurate masks compared to the baseline. For example, in Fig.~\ref{fig:qualitative_comparison} Contrastive-Lift consistently fails to predict the leg of the highlighted chair whereas our method consistently predicts the leg of the chair. Further, we provide more qualitative results in the supplementary material.



\begin{figure}[!t]
    \centering
    \includegraphics[width=\linewidth]{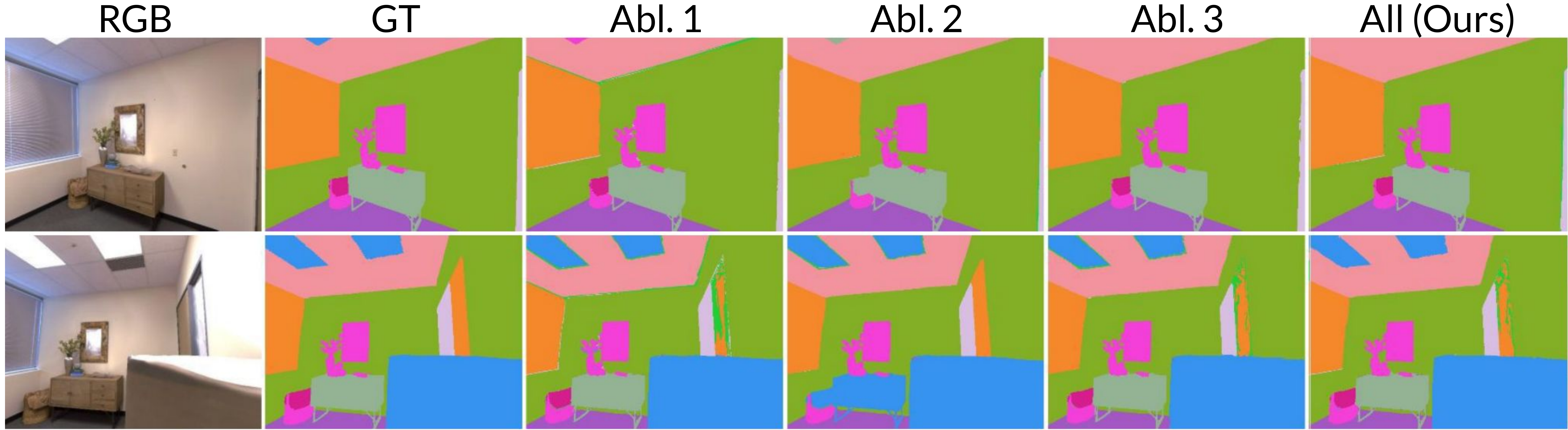}
    \caption{Qualitative results for different design choices discussed in ablation studies in Tab.~\ref{tab:ablations}.}
    \label{fig:ablations}
\end{figure}

\begin{table}[!tb]

\begin{adjustbox}{width=\linewidth}
\begin{tabular}{@{}c|ccccc@{}}
\toprule
\textbf{Contrastive-Loss (CL)} & \bluetick & \bluetick & \bluetick & \bluetick& \bluetick\\
\textbf{Triplet Loss} & \redcross & \redcross & \bluetick &  \bluetick & \bluetick\\
\textbf{3D Regularization} & \bluetick & \redcross & \redcross  & \bluetick & \bluetick \\
\textbf{MLP} & \redcross & \redcross & \bluetick  & \redcross & \bluetick\\ \midrule
 &
  \textbf{Abl. 1} &
  \textbf{Abl. 2} &
  \textbf{Abl. 3} &
    \textbf{w/o MLP} &
  \textbf{with MLP} \\ \midrule
\textbf{$PQ^{scene}$}   & 88.0  & 83.7                         & 89.0         & 88.0                 & 89.0 
   \\
\textbf{mIoU}& 94.4     & 91.8                      & 95.2       & 94.0                 & 95.4  
   \\ \bottomrule
\end{tabular}
\end{adjustbox}
\caption{ Effect of different loss functions. Abl.1 uses ``CL + 3D Regularization'' loss, Abl.2 uses only Contrastive loss and Abl.3 uses ``CL+Triplet loss with MLP''. We observe that a combination of Contrastive loss, Triplet loss and 3D neighborhood loss yields the best results for our method.}
\label{tab:ablations}
\end{table}

\begin{table}[!t]

\begin{adjustbox}{width=\linewidth}
\begin{tabular}{@{}cccccc@{}}
\toprule

 \textbf{Method} & \makecell{\textbf{3DGS Optimization} \\ \textbf{Time}} $\downarrow$ & \textbf{Clustering} $\downarrow$ & \makecell{\textbf{Total Training}\\ \textbf{Time}} $\downarrow$ & $\textbf{PQ}^{\textbf{scene}}$ $\uparrow$ & \textbf{IoU} $\uparrow$\\ \midrule
CL+3DGS & 42.2 m & 43.21 m & 85.41 m  & 94.6 & 97.4   \\ 
Ours & \textbf{42 m} & \textbf{0 m} & \textbf{42 m}  & 94 & 96.2   \\
\bottomrule
\end{tabular}
\end{adjustbox}
\caption{Quantitative comparison of our method with Contrastive-Lift (CL) + 3DGS}
\label{tab:discussion_hdbscan}
\end{table}

\begin{figure}[!t]
    \centering
    \includegraphics[width=\linewidth]{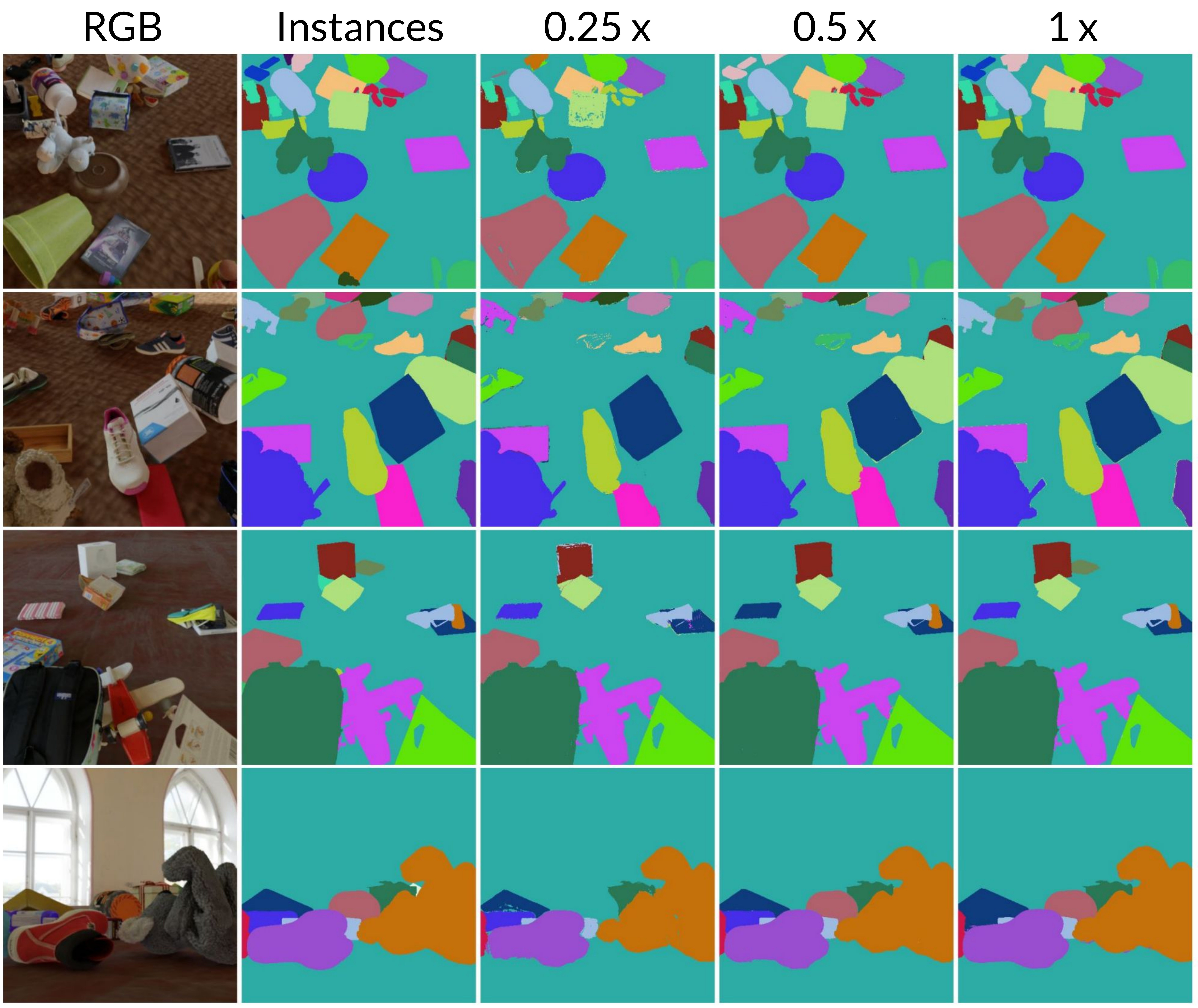}
    \caption{\textbf{Qualitative results on Messy-Rooms dataset by varying resolution of the segmentation masks.} We show results with $0.25\times$, $0.5\times$ and $1\times$(full) resolution}
    \label{fig:ablation_resolution}
\end{figure}

\begin{figure}[!t]
    \centering
    \includegraphics[width=\linewidth]{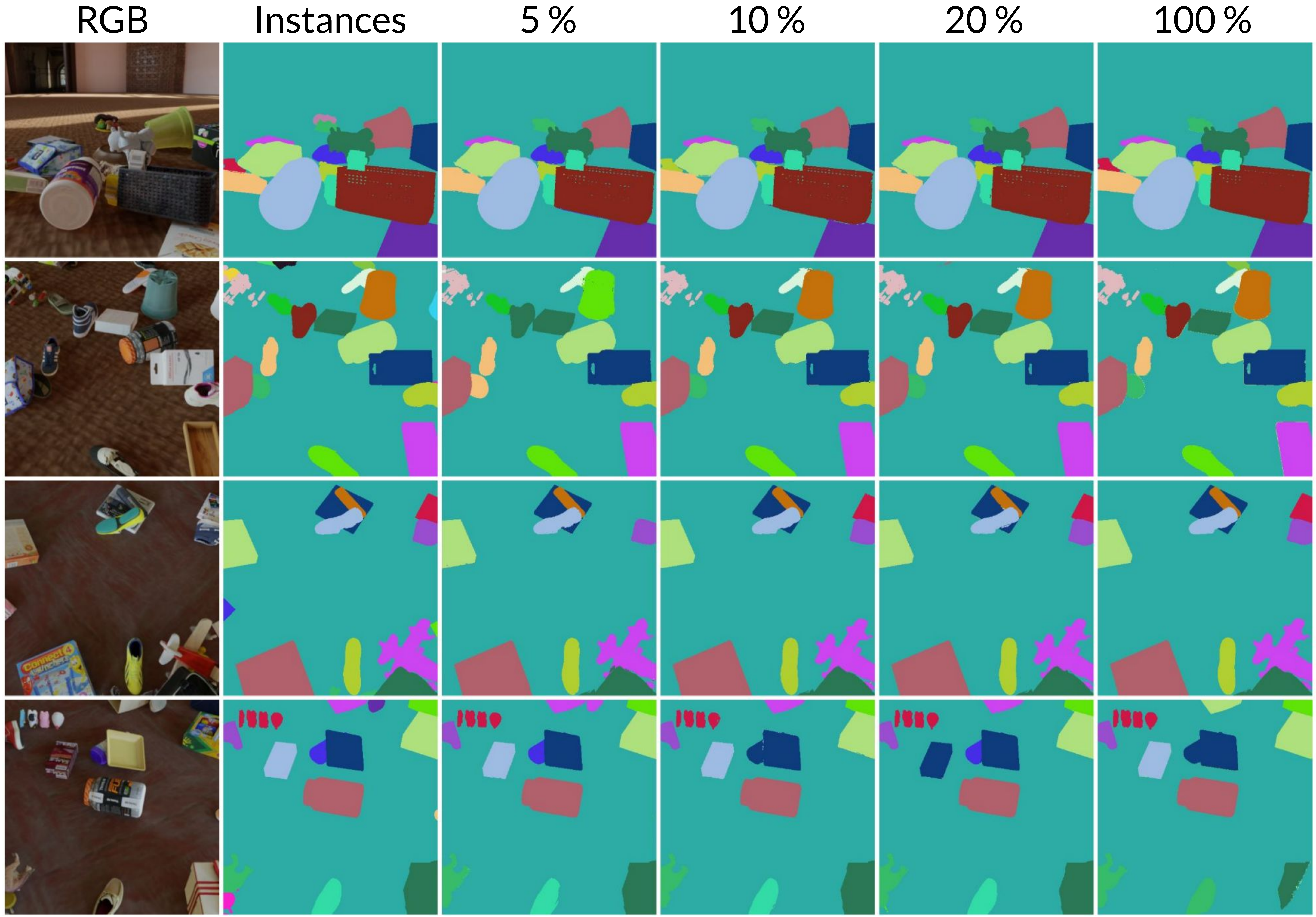}
    \caption{\textbf{Qualitative results on Messy-Rooms dataset by varying percentage of training data.} We show results with 5, 10, 20 and 100(full) \% training samples.}
    \label{fig:ablation_percentage}
\end{figure}

\noindent
\textbf{Time-performance comparison.}
Current methods for lifting 2D segmentation masks to 3D often suffer from slow training speeds due to complex, multi-stage pipelines. For instance, Panoptic-Lifting~\cite{siddiqui2023panoptic} relies on time-consuming linear assignment to resolve mask inconsistencies, while Contrastive-Lift~\cite{bhalgat2023contrastive} employs a computationally intensive clustering algorithm for post-processing. In contrast, our unified, end-to-end approach eliminates these costly intermediate steps, leading to significantly faster training, as shown in Tab.~\ref{tab:time_comparison}. To ensure a fair evaluation, we also compare against a stronger baseline, Contrastive-Lift-3DGS, which uses the same 3DGS representation as our method and $\mathcal{L}_{cluster}$ loss followed by HDBSCAN clustering. We observe that our approach is faster than this enhanced baseline as well (Tab.~\ref{tab:discussion_hdbscan}, with further analysis provided in the supplementary material.

\subsection{Ablations Study}

\noindent
\textbf{Contrastive Losses, 3D Loss} Our method uses a combination of 3D neighborhood loss, contrFastive loss and triplet loss  to optimize the feature field for instance segmentation purpose. We evaluate the impact of these losses. Tab.~\ref{tab:ablations} and Fig.~\ref{fig:ablations}  shows results when these losses are applied separately. We also observe that applying the triplet loss on MLP-projected embeddings yields better separation and improves the overall metrics. We observe that when we use a combination of the three losses, we obtain the best results.

\noindent
\textbf{Number of segmentation masks during training} 
We evaluated our method's robustness using a pre-trained 3DGS model on the Messy-Rooms dataset with only $5\%, 10\%$ and $20\%$ of segmentation masks. Fig.~\ref{fig:ablation_percentage} shows that the results achieve accuracy comparable to training with all masks, demonstrating the method's effectiveness with limited data. We provide implementation details for this ablation in the supplementary material.

\noindent
\textbf{Input resolution of the segmentation masks} This experiment investigates the impact of input segmentation mask resolution. We evaluate our method's ability to handle downscaled masks ($0.25\times$ and $0.5\times$) on four scenes from the Messy-Rooms dataset (Fig.~\ref{fig:ablation_resolution}). Visual inspection reveals no significant difference between results obtained with $0.5\times$and full-resolution masks. Training at a lower resolution will accelerate our method without compromising accuracy. We provide implementation details for this ablation in the supplementary material.

\section{Conclusion and Future Work}
In this work, we introduce~\methodname{}, a single-stage framework for lifting inconsistent 2D multi-view instance masks into a 3D representation. Unlike previous two-stage methods that rely on computationally expensive post-processing such as clustering, our approach directly decodes class labels from learned vector embeddings. These embeddings are optimized with a triplet loss function to reduce intra-class variance. Experiments demonstrate that UniC-Lift outperforms state-of-the-art methods while significantly reducing training time. This efficiency and accuracy enable practical applications such as interactive scene editing and mesh extraction, while its robust performance in complex, multi-object environments confirms its scalability. Future work focuses on extending~\methodname{} to dynamic scenes, large-scale unbounded scenes and hierarchical segmentation for a 3D representation.

\section{Acknowledgments}
This work was partly supported by KIAC, IISc. We thank Rishubh Parihar for reviewing the manuscript and providing insightful feedback.

\bibliography{aaai2026}

\clearpage

\appendix

\begin{figure*}[!t]
    \centering
    \includegraphics[width=\linewidth]{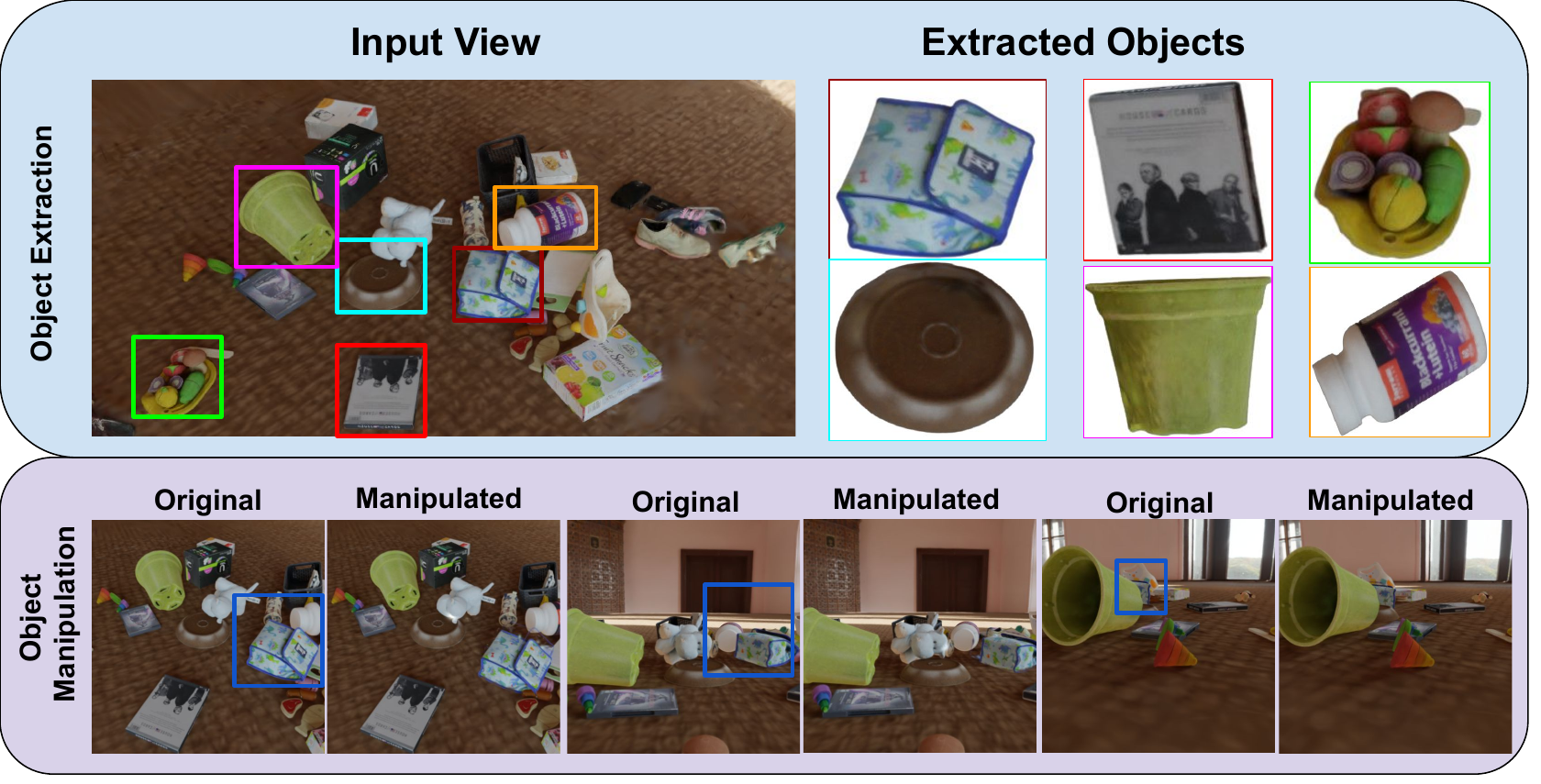}
    \caption{We show two downstream applications: Object Extraction and Object Manipulation. We can observe the effect of high-quality 3D segmentation. 3D objects of different shpaes and sizes are extracted very accurately. Further, we can also do editing operations like moving an object in the scene. We translate blue bag marked in blue box in the second row.  }
    \label{fig:applications}
\end{figure*}

\section{Additional Details}
\subsection{Discussion on conversion of embeddings to class label}
\label{sec:supp_discussion_conversion_of_embeddings_to_class_labels} 
\textit{Contrastive loss encourages similar objects to cluster in the feature space, but their exact positions are unpredictable. Clustering algorithms are typically needed to identify these clusters, which can be computationally expensive.} We restrict the range of rasterized vector embeddings $\mathcal{V}$ to $[0,1]$ by applying a sigmoid operation. Standard contrastive loss, applied within this constrained space,  drives embeddings toward specific corner points as shown in Fig.~\ref{fig:rebuttal-embedding-to-label}.  The resulting corner points act as a form of binary encoding, allowing for direct mapping of embeddings to unique class labels using a simple thresholding operation. To stabilize training and accelerate convergence, we incorporate additional loss terms to reduce variance in the embeddings. This ensures that different segments does not collapse to same integer. By eliminating the need for explicit clustering, this approach significantly improves inference speed and accuracy for instance and semantic segmentation tasks.

\subsection{What are two-stage methods?}
\label{sec:supp_discussion_two_stage_methods} 
\begin{figure}[!t]
  \centering
  \includegraphics[width=\linewidth]{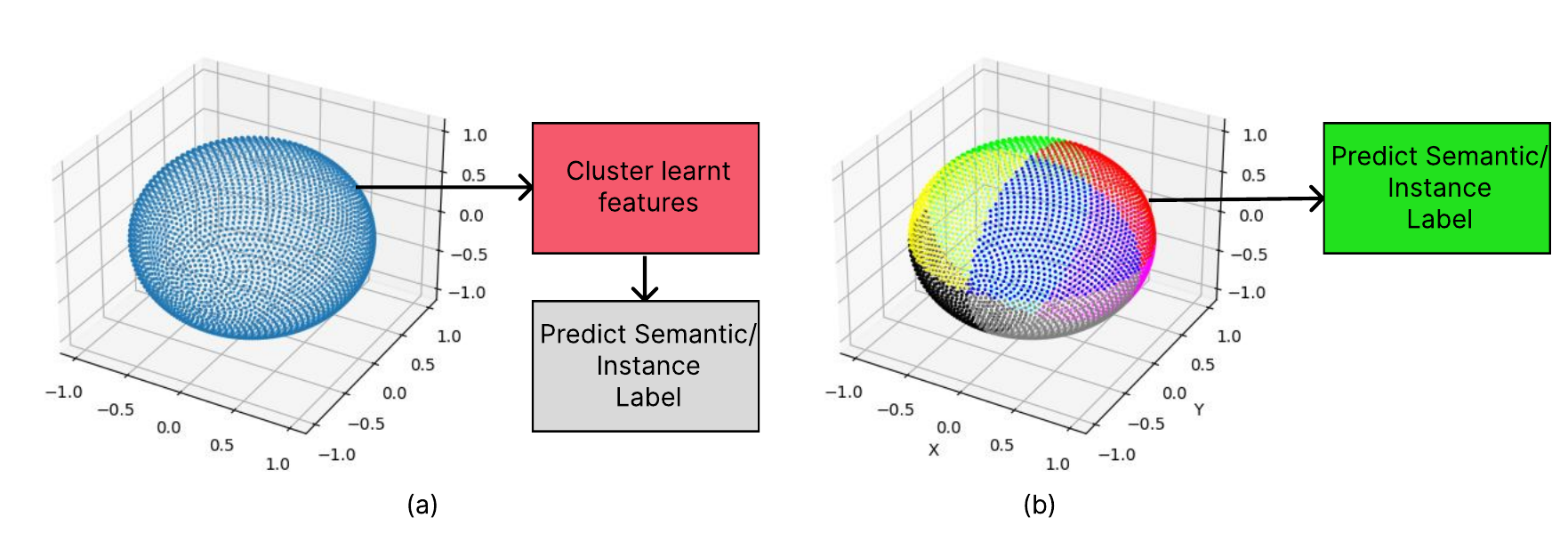} 
  \caption{(a) Methods such as Contrastive-Lift~\cite{bhalgat2023contrastive} learn the vector embeddings using a contrastive loss but rely on an off-the-shelf clustering algorithm (denoted in red) to find the centroids to predict labels. (b) Whereas we propose a single-stage optimization which directly predicts the segmentation label making our method faster.}
  \label{fig:supp_why_fig}
\end{figure}
Methods like Panoptic-Lift~\cite{fu2022panoptic} use a pre-processing (linear assignment or mask association) step for matching the predictions from 3D representations with the inconsistent 2D segmentation maps. Recent methods like Contrastive-Lift~\cite{bhalgat2023contrastive} utilize feature-fields and optimize them using contrastive learning and employ a clustering method like HDBSCAN to obtain the final consistent segmentation maps, which is hyperparameter sensitive. We illustrate this in Fig.~\ref{fig:supp_why_fig} where, for instance, labelling points on a sphere would first involve finding optimal vectors based on similarity. Then, a separate clustering step would be needed to assign labels to these vectors. Hence, these methods are two-stage methods. In contrast, our method is trained end-to-end and a unified representation for the lifting task, i.e. \textit{learned embeddings are directly decoded into class labels}, requiring no such additional pre- or post-processing. We have provided an exhaustive illustration in Fig.~\ref{fig:why_fig}.

\subsection{Why positive and negative samples are chosen from boundary?}
\label{sec:supp_discussion_hard_mining} 
Efficient learning relies on selecting triplets that provide meaningful gradients. \textit{Boundary triplets, where positive and negative distances are near the margin, yield non-zero gradients and help the network learn better features.} In contrast, \textit{random triplets often include "easy" cases with limited training value.} Focusing on boundary triplets provides more informative feedback, accelerating convergence during training. Based on our experiments, selecting boundary samples helped us achieve 94 $PQ^{scene}$ in $25k$ iterations. However, when using random triplet pairs, we required $50k$ iterations to attain the same quality.

\subsection{Comparison with Feature-3DGS}
\label{sec:supp_comparison_feature3dgs} 
\begin{figure}[!t]
    \centering
        \caption{Inference from Feature-3DGS~\cite{zhou2023feature} on MessyRooms~\cite{bhalgat2023contrastive} scene}
        \vspace{-4mm}
    \includegraphics[width=\linewidth]{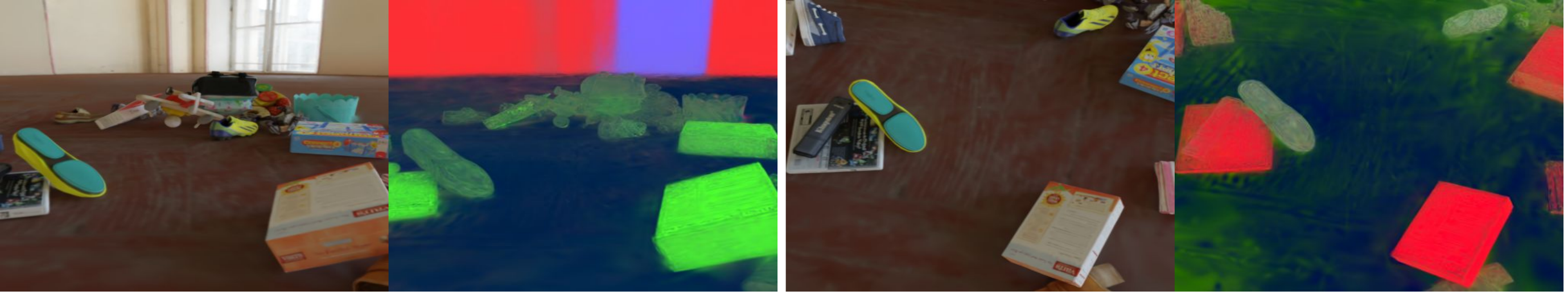}
    \vspace{-4mm}

    \label{fig:supp_feature_3dgs}
\end{figure}
\noindent 
Fig.~\ref{fig:supp_feature_3dgs} shows rendered feature fields from Feature-3DGS~\cite{zhou2023feature} for adjacent views of the ``Old Room'' scene (25 objects) from the Messy Rooms~\cite{bhalgat2023contrastive} dataset. Despite its dimensionality reduction capabilities, rendered feature fields exhibit significant multi-view inconsistencies. Hence, it is not suitable for 3D instance/panoptic segmentation tasks.

\section{Additional Results}
\label{sec:addition_results}

\subsection{Discussion on post-processing step in Contrastive-Lift}
\label{sec:discussion_hdbscan}

To validate the effectiveness of our approach, we replaced Contrastive-Lift’s backbone with 3DGS. After optimizing the feature embeddings, we applied the HDBSCAN algorithm to cluster them and generate class labels. Tab.~\ref{tab:discussion_hdbscan} shows the performance time and quantitative metrics on ``room\_1'' scene from Replica dataset on a single NVIDIA 3090 GPU. Notably, our method eliminates the need for post-processing clustering while maintaining comparable quantitative results. 

\subsection{Discussion on 3D Neighbourhood Regularization loss ($\mathcal{L}_{3d}$)}
\begin{table}[!t]
\centering
\begin{adjustbox}{width=0.7\linewidth}
\begin{tabular}{@{}c|cccc@{}}
\toprule
 Iterations & \textbf{7k} & \textbf{12k} & \textbf{15k (Ours)} & \textbf{20k}  \\ \midrule
 \textbf{$PQ_{scene}$}  & 60.5   &  65.6   &  65.2          &  64.6   \\
\textbf{mIoU}  &  71.3  &  77.8   &    74.5        &   75.0  \\ \bottomrule
\end{tabular}
\end{adjustbox}
\caption{\textbf{Ablation study on 3D Neighbourhood Regularization.} We observe that activating this loss at $15k$ balances accuracy and efficiency, making it the optimal choice.}
\label{tab:3d_neighbourhood_reg}
\end{table}
We justify the choice of 15k iterations for activating $\mathcal{L}_{3d}$ through an ablation study on another Replica3D scene in Tab.~\ref{tab:3d_neighbourhood_reg}, where it is enabled at 7k, 12k, and 20k iterations. where it is enabled at 7k, 12k, and 20k iterations. Activating too early (7k) or too late (20k) degrades accuracy. 12k yields comparable metrics, it increases overall runtime. Hence, activating at 15k balances accuracy and efficiency, making it the optimal choice. 

\subsection{Training for ablation with number of segmentation masks}
\label{sec:supp_abl_num_segmentation_maps}
In this ablation, we run only on $5\%, 10\%$ and $20\%$ of the segmentation masks. We utilize the entirety of the training RGB views for novel-view synthesis. Segmentation losses are then applied when a segmentation mask is available for a training view. We clarify that we do not use less RGB training views. 

\subsection{Training for ablation with resolution of the segmentation mask}
\label{sec:supp_abl_seg_mask_resolution}
To assess the impact of input resolution, we performed an ablation study where both the input segmentation mask and RGB training views were downscaled. During rendering, however, the masks were rendered at the original full resolution. Our results demonstrate that training at lower resolution does not significantly affect the final output quality compared to training at full resolution.

\subsection{Qualitative Results}
\label{sec:supp_qualitative_results}
We now present additional qualitative results on Messy-Room and Replica3D dataset for our method in the following pages. Check the accuracy and high-quality of our method.

\section{Applications}
Manipulating or editing object in a 3D scene requires accurate segmentation of the instance of 3D object. With our method we get accurate instance segmentation masks. These masks can then be used to move the entire object in the scene. We show in Fig.~\ref{fig:applications} by moving a blue bag to a different location. Further, we can extract individual objects very accurately, as shown in the first row of Fig.~\ref{fig:applications}. 
\begin{figure*}
    \centering
    \includegraphics[width=0.83\linewidth]{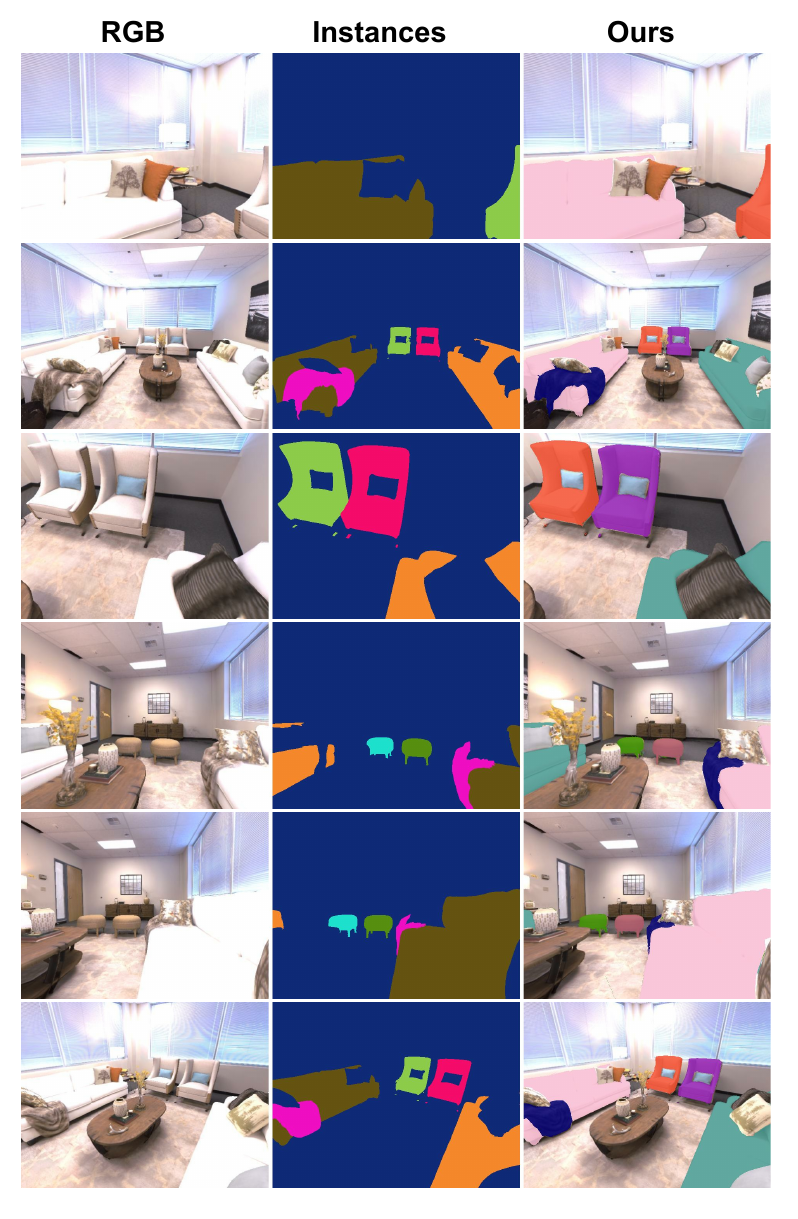}
    \caption{Room 0}
\end{figure*}

\begin{figure*}
    \centering
    \includegraphics[width=0.83\linewidth]{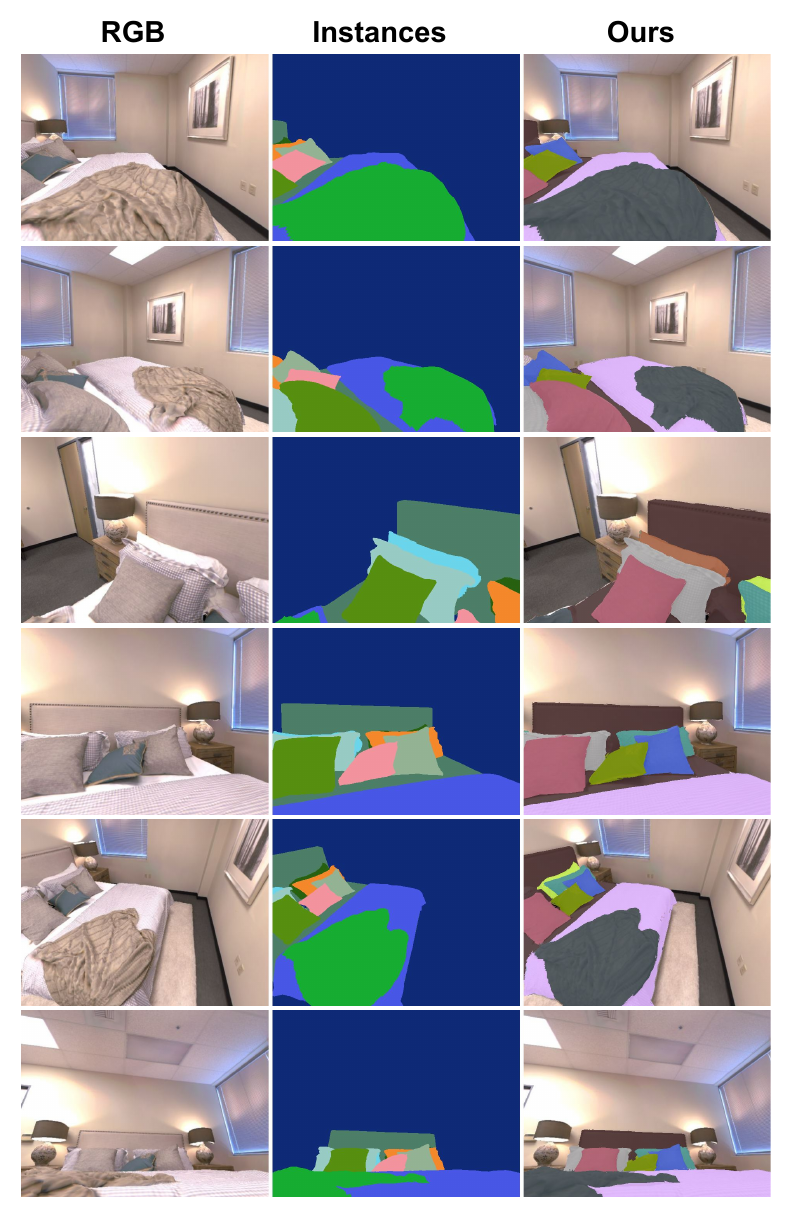}
    \caption{Room 1}
\end{figure*}

\begin{figure*}
    \centering
    \includegraphics[width=0.83\linewidth]{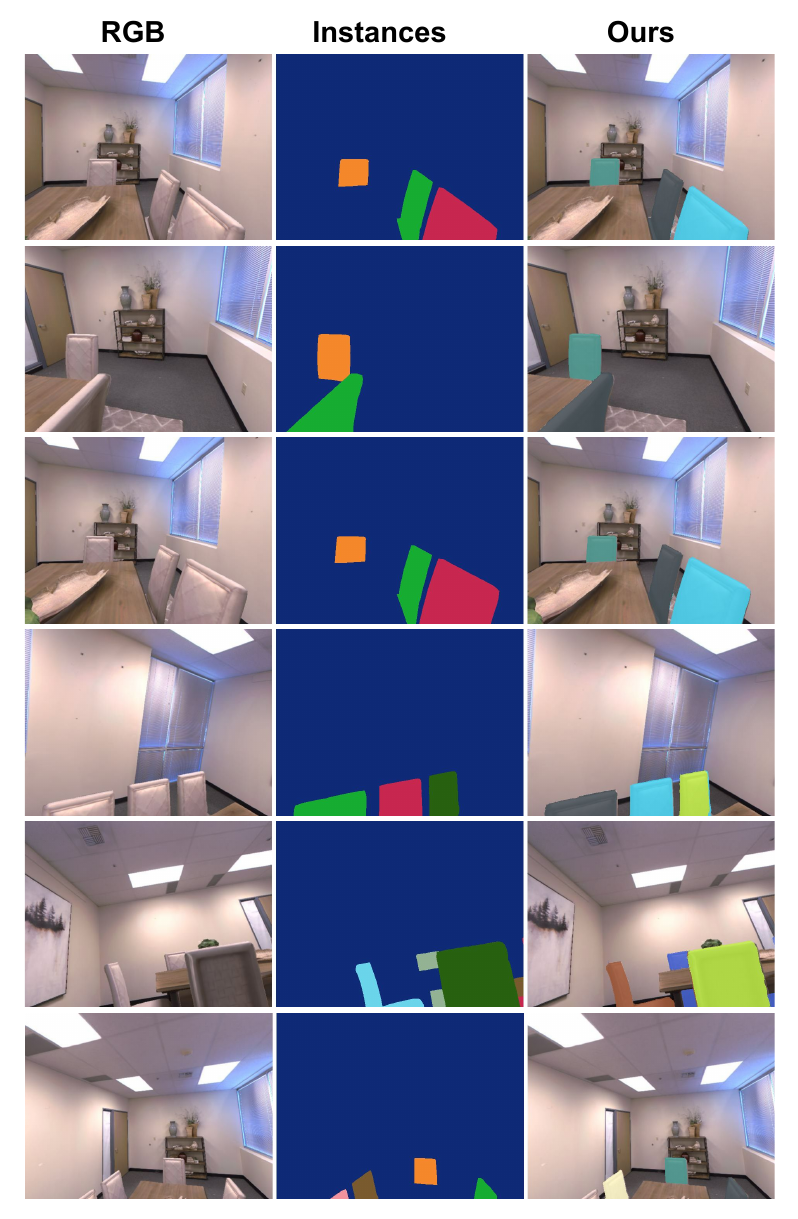}
    \caption{Room 2}
\end{figure*}

\begin{figure*}
    \centering
    \includegraphics[width=0.83\linewidth]{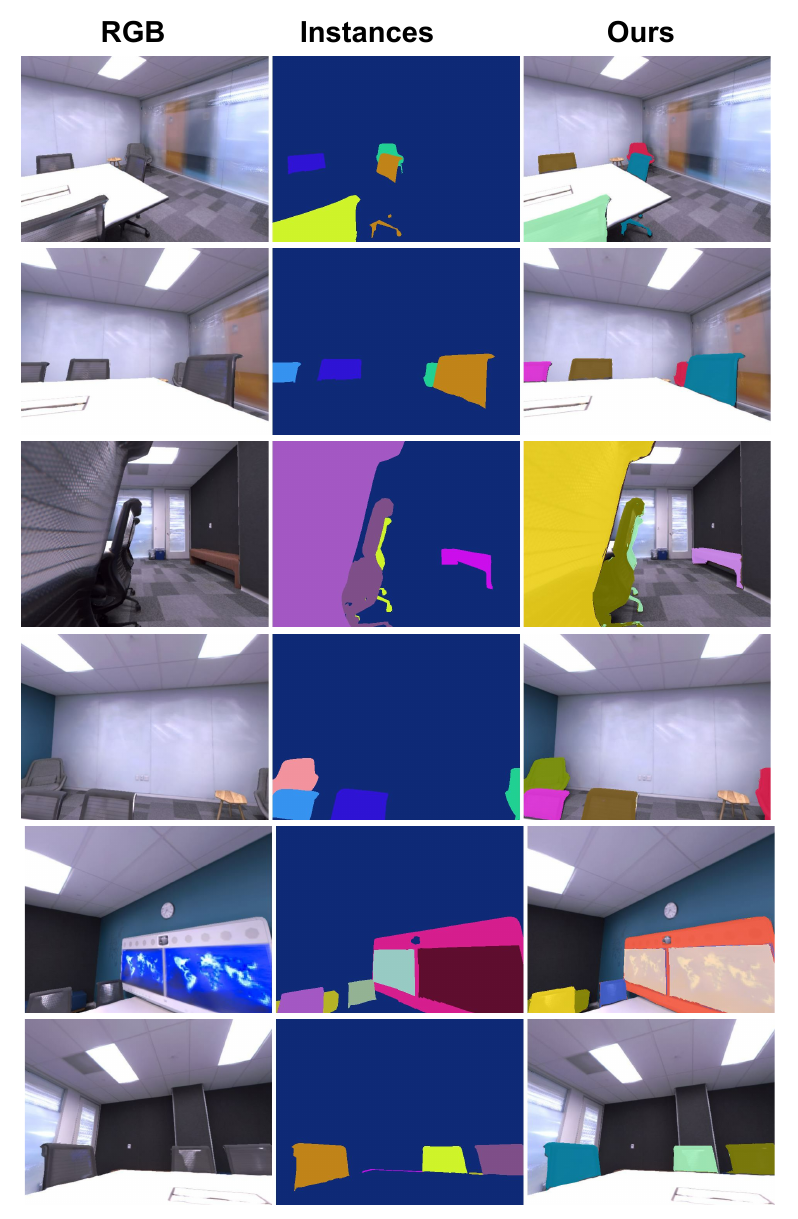}
    \caption{Office 4}
\end{figure*}

\begin{figure*}
    \centering
    \includegraphics[width=0.83\linewidth]{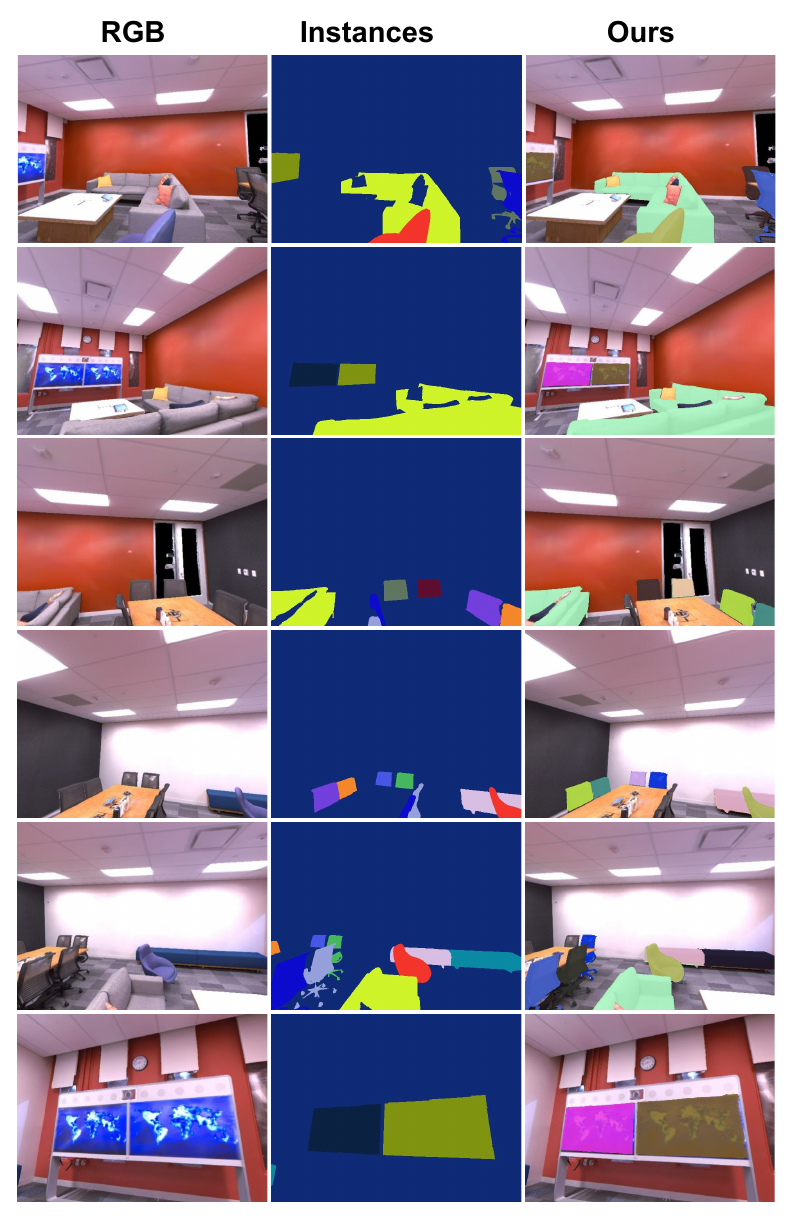}
    \caption{Office 3}
\end{figure*}

\begin{figure*}
    \centering
    \includegraphics[width=0.83\linewidth]{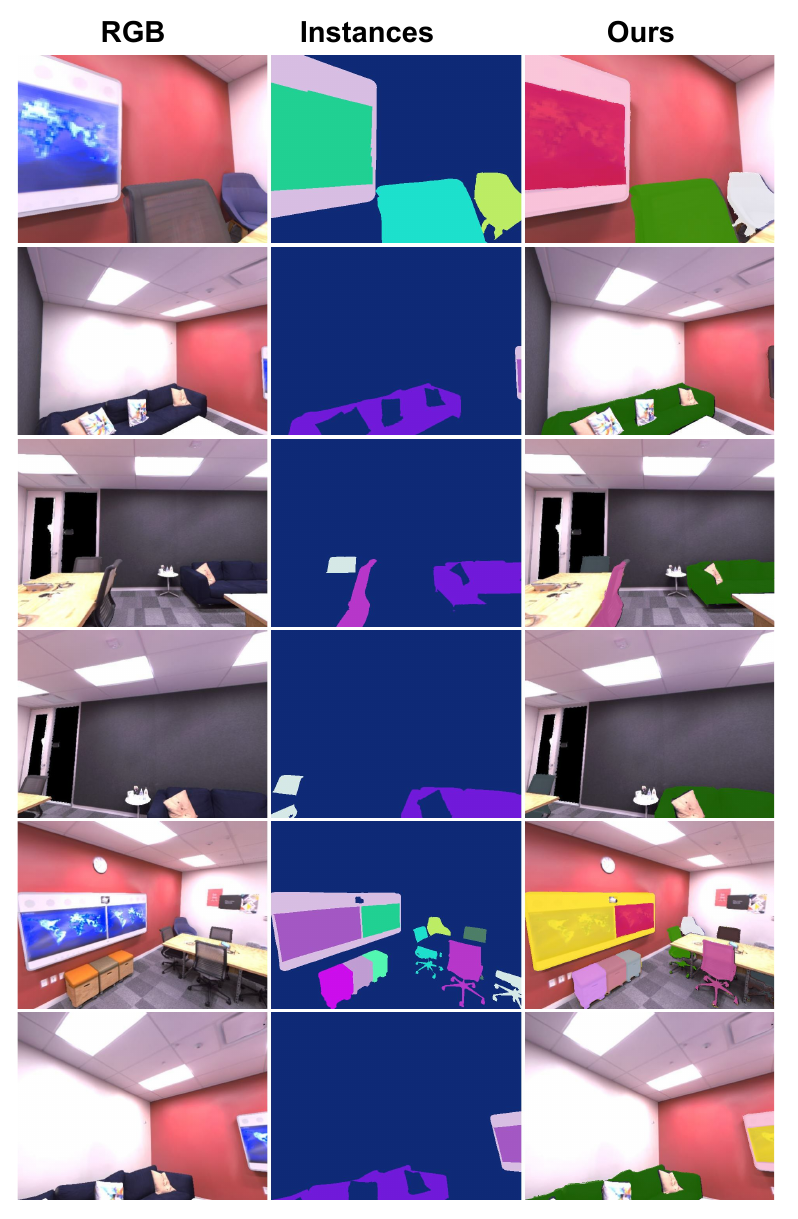}
    \caption{Office 2}
\end{figure*}

\begin{figure*}
    \centering
    \includegraphics[width=0.83\linewidth]{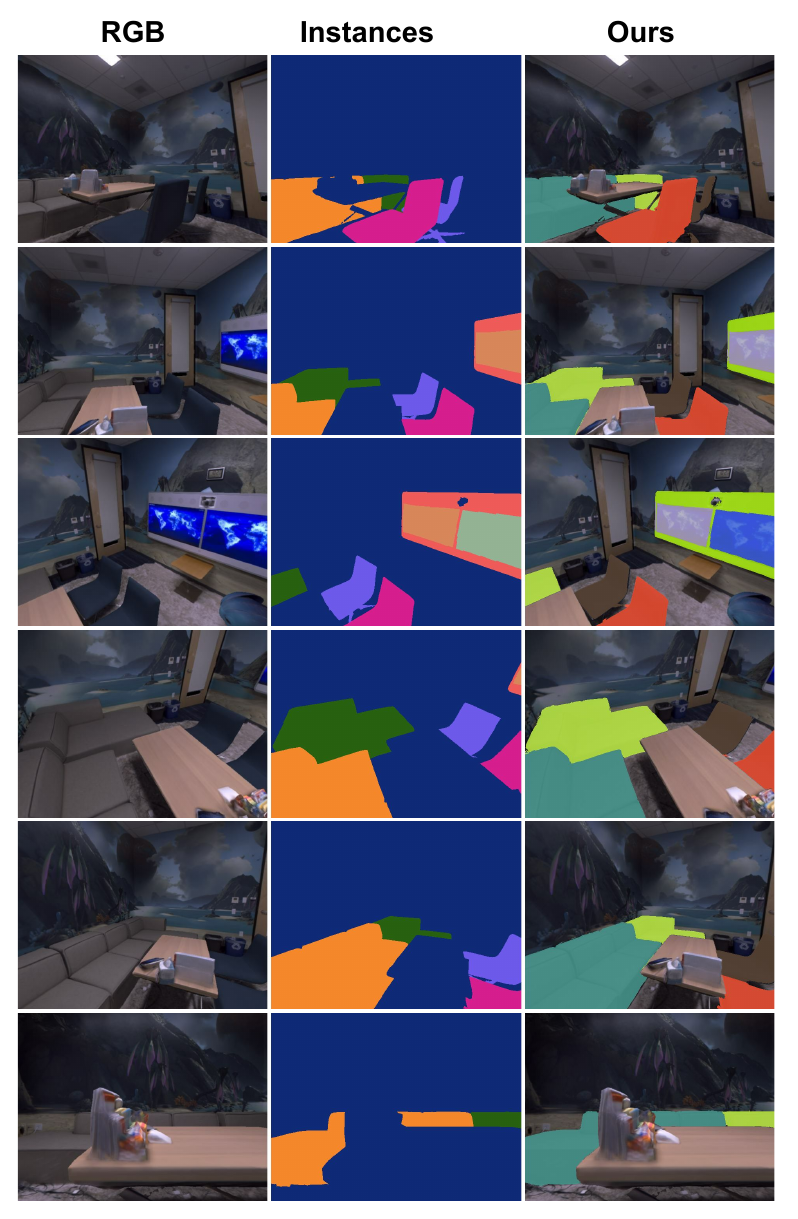}
    \caption{Office 1}
\end{figure*}

\begin{figure*}
    \centering
    \includegraphics[width=0.73\linewidth]{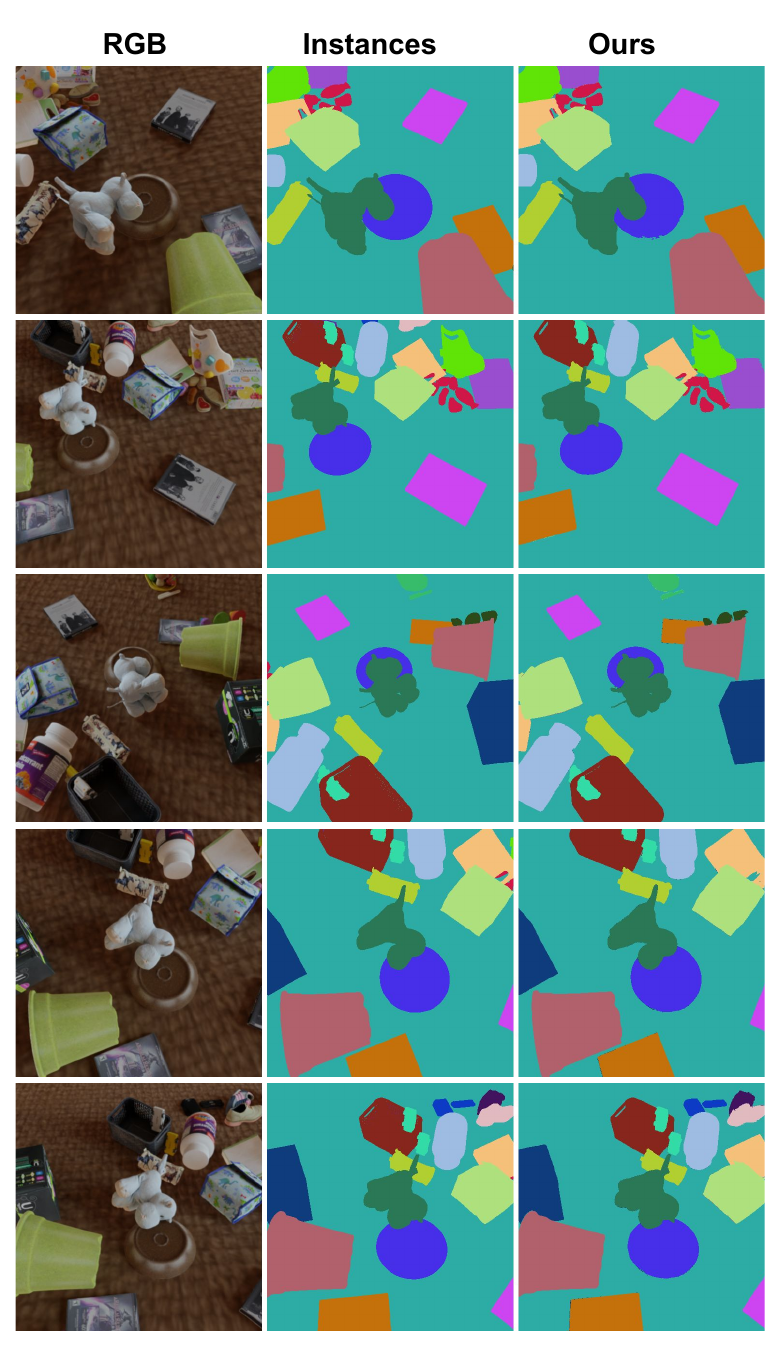}
    \caption{Large Corridor: 25 objects}
\end{figure*}

\begin{figure*}
    \centering
    \includegraphics[width=0.73\linewidth]{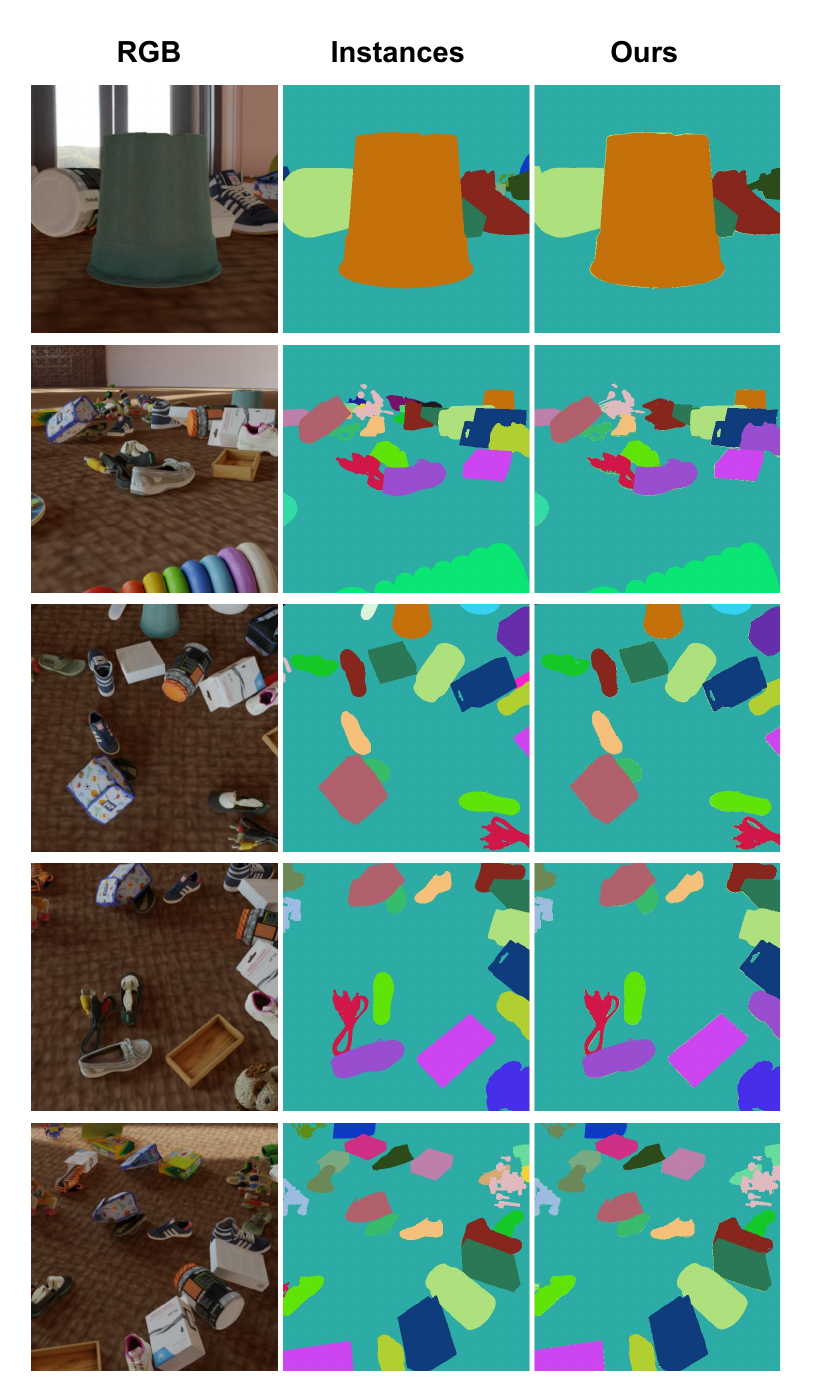}
    \caption{Large Corridor: 50 objects}
\end{figure*}

\begin{figure*}
    \centering
    \includegraphics[width=0.73\linewidth]{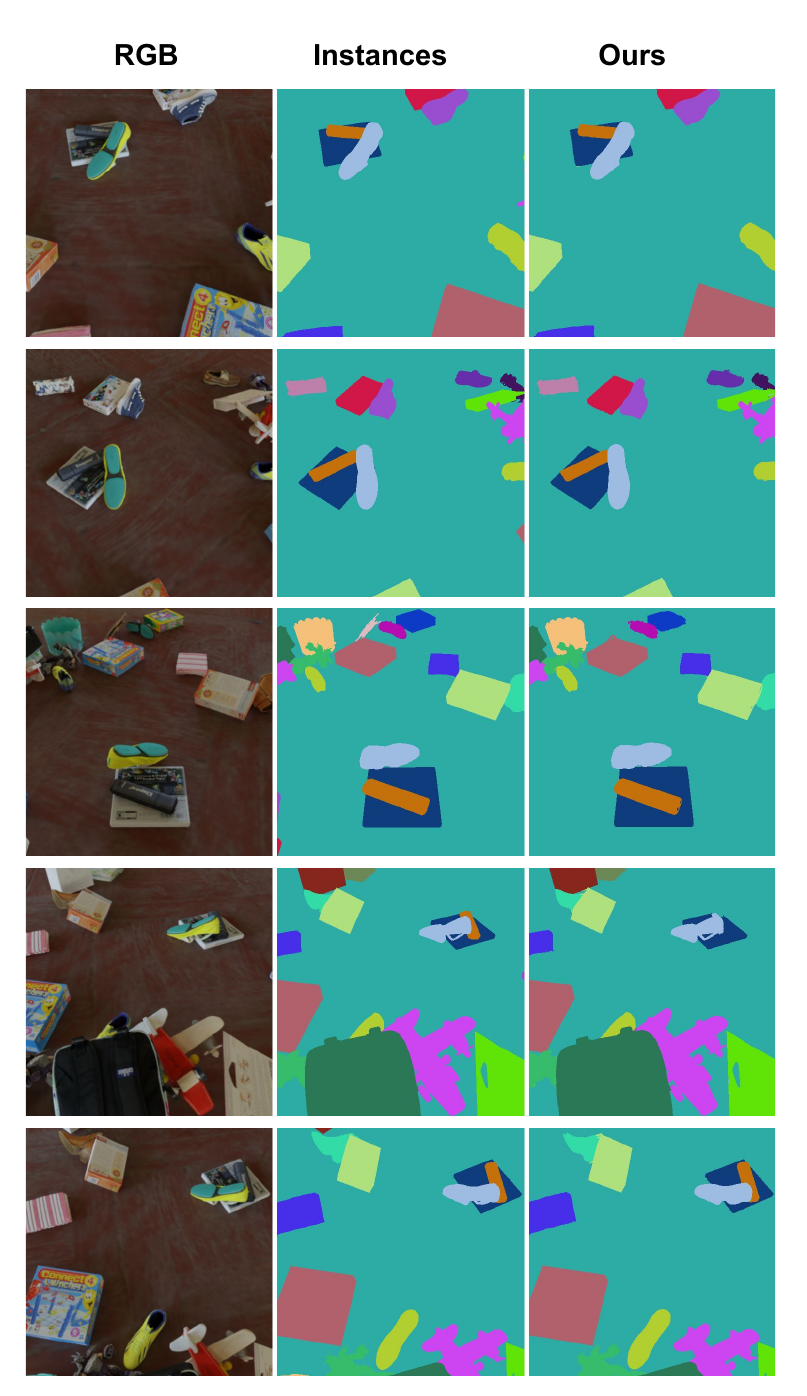}
    \caption{Old Room: 25 objects}
\end{figure*}

\begin{figure*}
    \centering
    \includegraphics[width=0.71\linewidth]{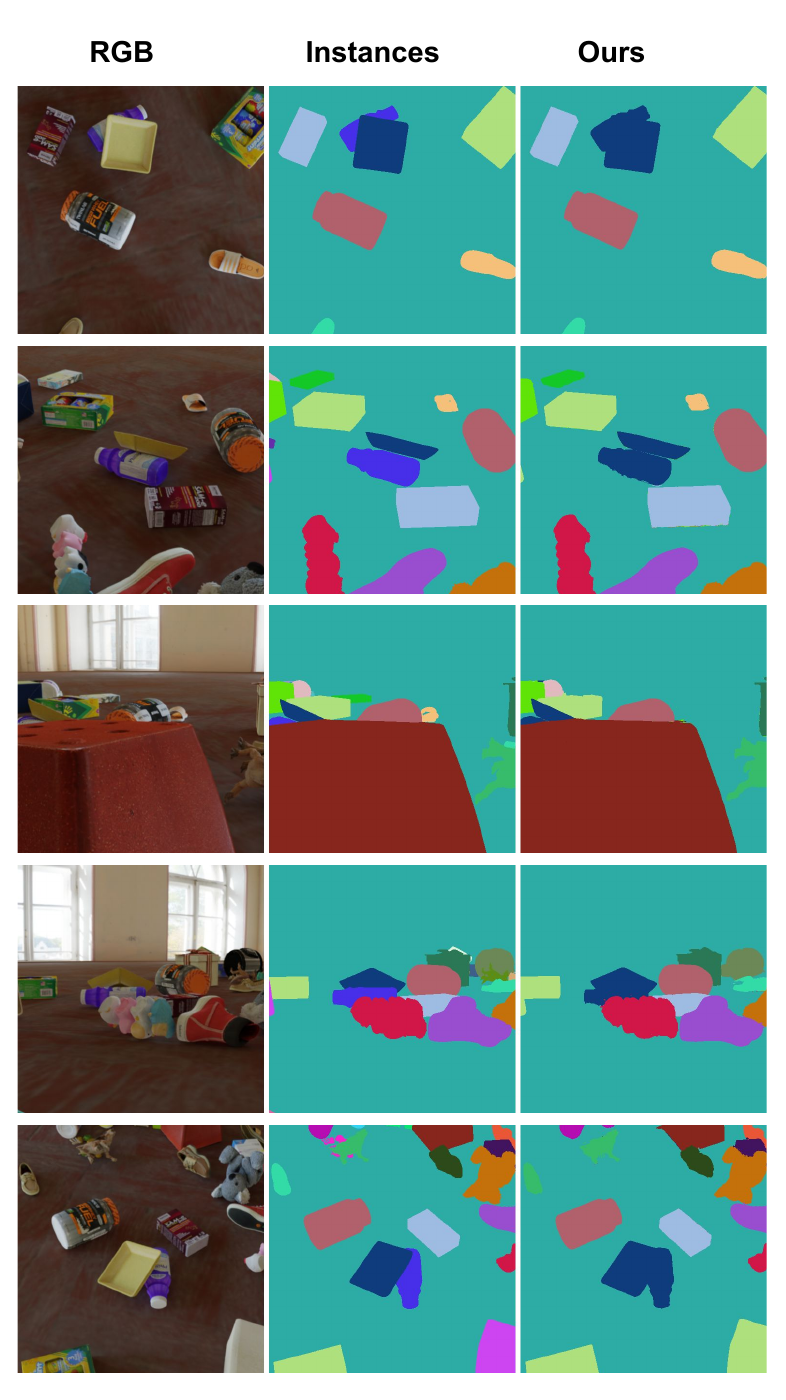}
    \caption{Old Room: 50 objects}
\end{figure*}

\end{document}